\documentclass[runningheads]{llncs}

\usepackage{eccv}

\usepackage{eccvabbrv}

\usepackage{graphicx}
\usepackage{booktabs}
\usepackage{hyperref}
\usepackage{url}
\usepackage{float}
\usepackage{multirow}  
\usepackage{tabularx}
\usepackage{enumitem}
\usepackage{wrapfig}
\usepackage{bbm} 
\usepackage{xfrac}
\usepackage[table]{xcolor}
\usepackage{xcolor}
\usepackage{pifont}
\usepackage{makecell}
\definecolor{checkgreen}{RGB}{29, 177, 134} 
\definecolor{crossred}{RGB}{194, 59, 53}    

\newcommand{\cmark}{\textcolor{checkgreen}{\ding{51}}}
\newcommand{\xmark}{\textcolor{crossred}{\ding{55}}}

\usepackage[accsupp]{axessibility}  

\usepackage{xcolor}
\definecolor{linkcolor}{RGB}{0, 102, 204}   
\hypersetup{urlcolor=linkcolor, colorlinks=true}

\newif\ifdrafting
\draftingtrue %
\ifdrafting
    \newcommand{\todotxt}[1]{{\leavevmode\color[rgb]{1,0,0}[TODO: #1]}}
    \newcommand{\ds}[1]{{\leavevmode\color[rgb]{0.8,0,0.8}[Deqing: #1]}}
    \newcommand{\cih}[1]{{\leavevmode\color[rgb]{0,0.5,0}[Charles: #1]}}
    \newcommand{\jh}[1]{{\leavevmode\color[rgb]{0.8, 0.2, 0}[Junhwa: #1]}}
    \newcommand{\leslie}[1]{{\leavevmode\color[rgb]{0.7, 0, 0.7}[Leslie: #1]}}
\else
    \newcommand{\todotxt}[1]{}
    \newcommand{\ds}[1]{}
    \newcommand{\cih}[1]{}
    \newcommand{\jh}[1]{}
    \newcommand{\leslie}[1]{}
\fi

\usepackage{hyperref}

\usepackage{orcidlink}

\newcommand{\myparagraph}[1]{\noindent\textbf{#1}}
\begin{document}

\newcommand{\ourmethod}{GeCo}
\title{\ourmethod{}: Evaluating Geometric Consistency for Video Generation via Motion and Structure} 

\titlerunning{\ourmethod{}: Evaluating Geometric Consistency for Video Generation}

\author{Leslie Gu$^{1}$ \quad
    Junhwa Hur$^{2}$ \quad
    Charles Herrmann$^{2}$ \quad
    Fangneng Zhan$^{3}$ \\
    Todd Zickler$^{1}$ \quad
    Deqing Sun$^{2}$ \quad
    Hanspeter Pfister$^{1}$}

\authorrunning{L.~Gu et al.}
\institute{$^{1}$Harvard University \qquad
    $^{2}$Google DeepMind
    \qquad
    $^{3}$MIT}
\maketitle
\begin{center}
\vspace{-1.5em}\textbf{\url{https://GeCo-GeoConsistency.github.io}}
\end{center}
\vspace{-2.5em}
\begin{figure}[h!]
\centering
\includegraphics[width=\textwidth]{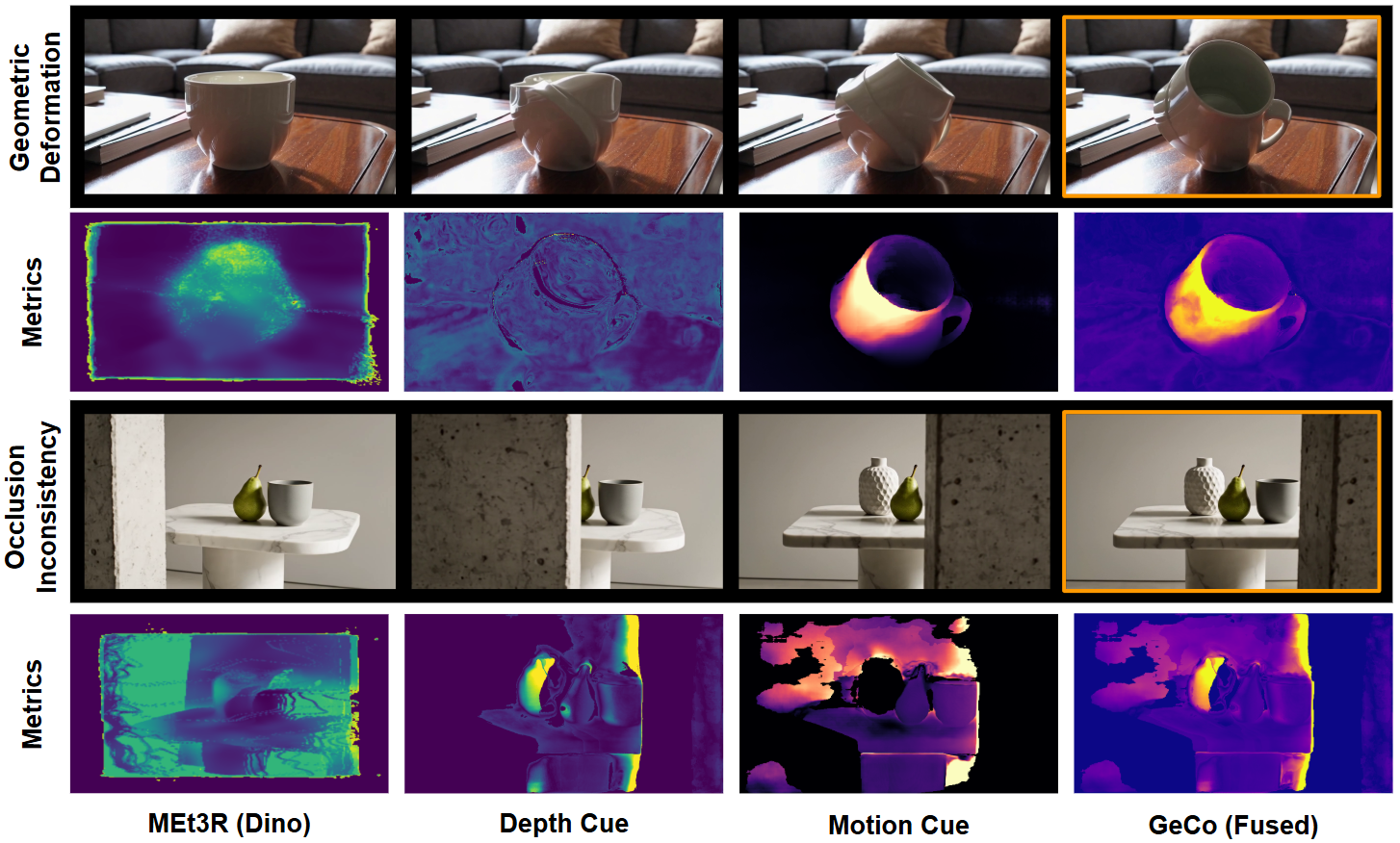}
\vspace{-1.5em}
\caption{
\textbf{Artifact videos}: a cup undergoes geometric deformation; a vase is hallucinated behind the pear after occlusion; orange boxes mark the target frames.
\textbf{Comparison of evaluation methods}: MEt3R~\cite{Asim2025MEt3RMM} fails to pinpoint geometric deformation and gives a vague signal on occlusion inconsistency. Depth cue captures occlusion inconsistency but does not localize surface deformation. Motion cue highlights non-rigid deformation artifacts, but provides no signal in occluded regions and thus misses occlusion inconsistency artifacts. \ourmethod{} (Fused) integrates motion and depth cues to accurately localize both types of artifacts in a single interpretable map.
}
\label{fig:teaser}
\end{figure}
\vspace{-3em}

\begin{abstract}
Visual generation models can produce photorealistic videos yet violate multiview geometry, exhibiting non-rigid deformations and occlusion inconsistencies (e.g., hallucinated content in disoccluded regions).
These failures hinder downstream applications such as video world model development and 3D asset creation, and are poorly captured by existing metrics.
We introduce \ourmethod{}, a geometry-grounded metric for jointly detecting geometric deformation and occlusion-inconsistency artifacts in static scenes. 
By fusing residual motion and depth priors, \ourmethod{} produces interpretable, dense consistency maps that localize these artifacts. 
Using \ourmethod{}, we systematically benchmark recent video generation models, revealing common geometric failure modes. 
We further show that \ourmethod{} provides an actionable signal by applying it as a training-free guidance loss that substantially reduces geometric artifacts during generation.

\end{abstract}
\section{Introduction}
\label{sec:intro}
Recent generative video models have achieved remarkable photorealism~\cite{brooks2024sora,veo3techreport,kong2024hunyuanvideo,yang2024cogvideox} and are increasingly used in applications that require geometric consistency across viewpoints, such as video world models~\cite{zhang2025matrixgame,he2025matrix,hyworld2025,worldplay2025,mao2025yume,RelicWorldModel2025} and 3D asset creation~\cite{han2024vfusion3d,voleti2024sv3d,VideoScene,agarwal2025cosmos}.
However, even when genearting nominally static scenes, these models frequently produce perceptually salient artifacts that violate multiview geometry.

We identify two primary, complementary geometric failure modes in static scenes: (i) Geometric deformation: co-visible rigid surfaces stretch, bend, or ``melt'' under camera motion. (ii) Occlusion inconsistency: previously occluded objects reappear altered, or disoccluded regions suffer from hallucinatory content shifts, violating object permanence.
Because these artifacts arise in different visibility regimes---co-visible surfaces versus newly revealed regions---reliable geometric evaluation must detect both jointly.

Yet existing metrics capture only part of this picture. 
General video benchmarks~\cite{Huang2023VBenchCB,Huang2024VBenchCA} lack explicit 3D geometric evaluation. Epipolar constraints~\cite{Yu2023PhotoconsistentNVS,agarwal2025cosmos} and reprojection-based world-model suites~\cite{duan2025worldscore} rely on sparse point matching, which inherently fails in occluded regions~\cite{Caron2021EmergingPI}---precisely where occlusion inconsistency manifests. Conversely, dense multi-view metrics based on 3D warped semantic features~\cite{Asim2025MEt3RMM} expose disocclusion errors but lack the geometric precision to penalize subtle surface deformations. As summarized in Table~\ref{tab:related_work}, no existing method offers a dense, interpretable diagnostic for both artifact types.


To bridge this gap, we propose \ourmethod{}, a dense, geometry-grounded metric that jointly localizes geometric deformation and occlusion inconsistency. \ourmethod{} leverages feed-forward estimators for optical flow, depth, and camera pose to compute two complementary signals: a \textit{motion cue} measuring flow-based rigidity (for deformation), and a \textit{structure cue} tracking depth reprojection consistency (for occlusion errors). \ourmethod{} fuses the cues into a unified, scale-invariant, per-pixel error map. Crucially, because all components are differentiable, \ourmethod{} serves as both an interpretable evaluation metric and an inference-time guidance signal.


We validate \ourmethod{} using two controlled datasets: WarpBench (thin-plate-spline warps isolating geometric deformation) and OccluBench (composited edits simulating occlusion inconsistencies). Experiments confirm our motion and structure cues are highly complementary, robust to off-the-shelf estimator noise, and their fusion substantially outperforms semantic baselines~\cite{Asim2025MEt3RMM}. We demonstrate \ourmethod{}'s versatility in two scenarios: first, as a metric benchmarking state-of-the-art video models across diverse static scenes (\eg, object-centric, indoor, outdoor), revealing consistent geometric failure patterns; second, as a training-free inference guidance objective that suppresses spurious motion and significantly improves both geometric consistency and downstream 3D reconstruction.



In summary, our contributions are:
\begin{itemize}
    \item \ourmethod{}: A differentiable, dense metric that jointly detects geometric deformation and occlusion inconsistency, producing interpretable error maps.
    \item \ourmethod{}-Eval: A geometric consistency evaluation suite for static scenes to benchmark video generation models, along with a systematic comparison of state-of-the-art models.
    \item Inference-Time Guidance: A training-free inference-time optimization that uses \ourmethod{} as a guidance objective during sampling to reduce geometric artifacts and improve consistency.
    \item Failure-Mode Analysis: An analysis of spurious motion artifacts in recent text-to-video models, showing that \ourmethod{}-based guidance reduces erroneous motion in otherwise static regions.
\end{itemize}

\begin{table*}[tb]
\centering
\caption{\textbf{Comparison of GeCo and existing metrics.} We factor metrics into a two-stage pipeline: (i) correspondence matching and (ii) cue comparison. We report whether each method yields a dense, interpretable consistency map and whether it is diagnostic for subtle geometric deformation and occlusion inconsistency artifacts.}
\vspace{-0.5em}
\label{tab:related_work}
\resizebox{\textwidth}{!}{%
\begin{tabular}{lllccc}
\toprule
\textbf{Method} &
\textbf{Correspondence} &
\textbf{Cue} &
\textbf{Dense} &
\textbf{Deform.} &
\textbf{Occlusion} \\
\midrule
\textbf{TSED}~\cite{Yu2023PhotoconsistentNVS} &
SIFT~\cite{Lowe1999ObjectRF} &
Symmetric epipolar distance~\cite{HartleyZisserman2003MVG} &
\xmark & \cmark & \xmark \\
\textbf{Cosmos}~\cite{agarwal2025cosmos} &
SuperPoint~\cite{DeTone2017SuperPointSI} \& LightGlue~\cite{Lindenberger2023LightGlueLF} &
Sampson error~\cite{Sampson1982FittingCS} &
\xmark & \cmark & \xmark \\
\textbf{VBench}~\cite{Huang2023VBenchCB,Huang2024VBenchCA} &
None (frame-wise) &
CLIP feature~\cite{Radford2021LearningTV} &
\xmark & \xmark & \xmark \\
\textbf{MEt3R}~\cite{Asim2025MEt3RMM} &
3D warping (DUSt3R~\cite{wang2024dust3r}) &
DINO feature~\cite{Caron2021EmergingPI} &
\cmark & \xmark & \cmark \\
\textbf{WorldScore}~\cite{duan2025worldscore} &
Learnt flow (DROID-SLAM~\cite{teed2021droid}) &
Reprojection error &
\cmark & \cmark & \xmark \\
\midrule
\textbf{GeCo - Motion} &
Optical flow (UFM~\cite{zhang2025ufm}) &
Residual motion &
\cmark & \cmark & \xmark \\
\textbf{GeCo - Structure} &
3D warping (VGGT~\cite{wang2025vggt}) &
Reprojection error &
\cmark & \xmark & \cmark \\
\textbf{GeCo - Fused} &
Flow \& 3D warping &
Fused &
\cmark & \cmark & \cmark \\
\bottomrule
\end{tabular}%
}
\vspace{-1.5em}
\end{table*}

\section{Related Works}
\label{sec:related}

\subsection{Video Generation Benchmarks}
Comprehensive benchmarks such as VBench~\cite{Huang2023VBenchCB,Huang2024VBenchCA} and EvalCrafter~\cite{liu2024evalcrafter} evaluate video generation along perceptual axes (e.g., temporal coherence, text--video alignment) but do not explicitly target multi-view geometric consistency. Another line of work measures physical plausibility~\cite{meng2024PhyGenBench,bansal2024videophy,xue2025phyt2v}, focusing on realistic dynamics rather than static-scene geometry under camera motion. WorldScore~\cite{duan2025worldscore} includes a depth-reprojection-based 3D-consistency score, but remains limited under occlusion and therefore less sensitive to occlusion inconsistency. To fill this gap, we target geometric consistency in static scenes, evaluating both deformation and occlusion inconsistency.

\subsection{Geometric Consistency Metrics}\label{subsec:metrics}
Most geometric consistency metrics follow a two-stage pipeline: (i) establish inter-frame correspondences and (ii) compare cues on the matched content. Table~\ref{tab:related_work} summarizes representative designs in terms of dense interpretability, deformation sensitivity, and occlusion-inconsistency awareness.

\noindent\textbf{Correspondence matching.}
Correspondences are typically obtained via (a) sparse keypoints~\cite{Yu2023PhotoconsistentNVS,agarwal2025cosmos}, (b) dense flow~\cite{duan2025worldscore}, or (c) 3D reconstruction-based warping~\cite{Asim2025MEt3RMM} (\ie, reprojecting co-visible 3D points into the target view).
Keypoint- and flow-based pipelines are inherently undefined in occlusion regions where no reliable matches exist; these regions are therefore commonly masked, directly ignoring occlusion inconsistency artifacts.
Reconstruction-based warping is visibility-aware: it can identify co-visible pixels across frames, enabling comparison in disoccluded regions where hallucinated content may appear.

\noindent\textbf{Cue comparison.}
Given correspondences, methods compare either semantic features or geometric errors.
Semantic features (\eg, DINO~\cite{Caron2021EmergingPI}, CLIP~\cite{Radford2021LearningTV}) may under-penalize subtle geometric distortions when semantics are preserved.
Epipolar-constraint cues (\eg, symmetric epipolar distance~\cite{HartleyZisserman2003MVG}, Sampson error~\cite{Sampson1982FittingCS}) are less sensitive when epipolar lines have similar orientations, as the epipolar error is insensitive to correspondence errors parallel to the epipolar line~\cite{Yu2023PhotoconsistentNVS}.
Depth cues depend on the accuracy of the underlying depth model and, when paired with 3D warping, may under-penalize deformations that remain close in 3D (\ie, small depth residuals despite incorrect correspondences).

\noindent\textbf{Representative methods.}
TSED~\cite{Yu2023PhotoconsistentNVS} uses SIFT~\cite{Lowe1999ObjectRF} with thresholded symmetric epipolar distance~\cite{HartleyZisserman2003MVG}; Cosmos (Geometric Consistency)~\cite{agarwal2025cosmos} uses SuperPoint~\cite{DeTone2017SuperPointSI} and LightGlue~\cite{Lindenberger2023LightGlueLF} with Sampson error~\cite{Sampson1982FittingCS}; and VBench (Background Consistency)~\cite{Huang2023VBenchCB,Huang2024VBenchCA} compares global CLIP features~\cite{Radford2021LearningTV} without explicit correspondence.
MEt3R~\cite{Asim2025MEt3RMM} uses DUSt3R~\cite{wang2024dust3r} warping with DINO features~\cite{Caron2021EmergingPI}, while WorldScore (3D Consistency)~\cite{duan2025worldscore} uses DROID-SLAM~\cite{teed2021droid} to compute dense reprojection error from learned-flow correspondences.
Our method fuses dense flow~\cite{zhang2025ufm} and 3D warping~\cite{wang2025vggt} to produce dense, interpretable maps that capture both subtle deformation and occlusion inconsistency within a unified metric.


\subsection{Improving Geometric Consistency in Video Generation}
Training-based approaches improve 3D consistency through joint video-geometry estimation~\cite{huang2025jog3r,zhang2025world} or explicit 3D caches~\cite{ren2025gen3c}, but require large-scale 3D annotations (depth, pointmaps, camera poses) obtained from off-the-shelf estimators~\cite{wang2024dust3r,bhat2023zoedepth,yang2024depth,teed2021droid}, which are often costly to prepare.
Orthogonally, training-free approaches steer diffusion sampling at inference time without updating model parameters, using guided sampling~\cite{Bansal2023UniversalGF,han2024multistable}, structural control~\cite{zhang2305controlvideo,zhang2023adding}, or flow-based objectives~\cite{Nam2024OpticalFlowGP,geng2024motion} to impose motion or structural consistency.
In this work, we apply \ourmethod{} as training-free guidance at inference time, showcasing that it provides an effective signal for improving geometric consistency during sampling.

\begin{figure*}[tb]
    \centering
    \includegraphics[width=\textwidth]{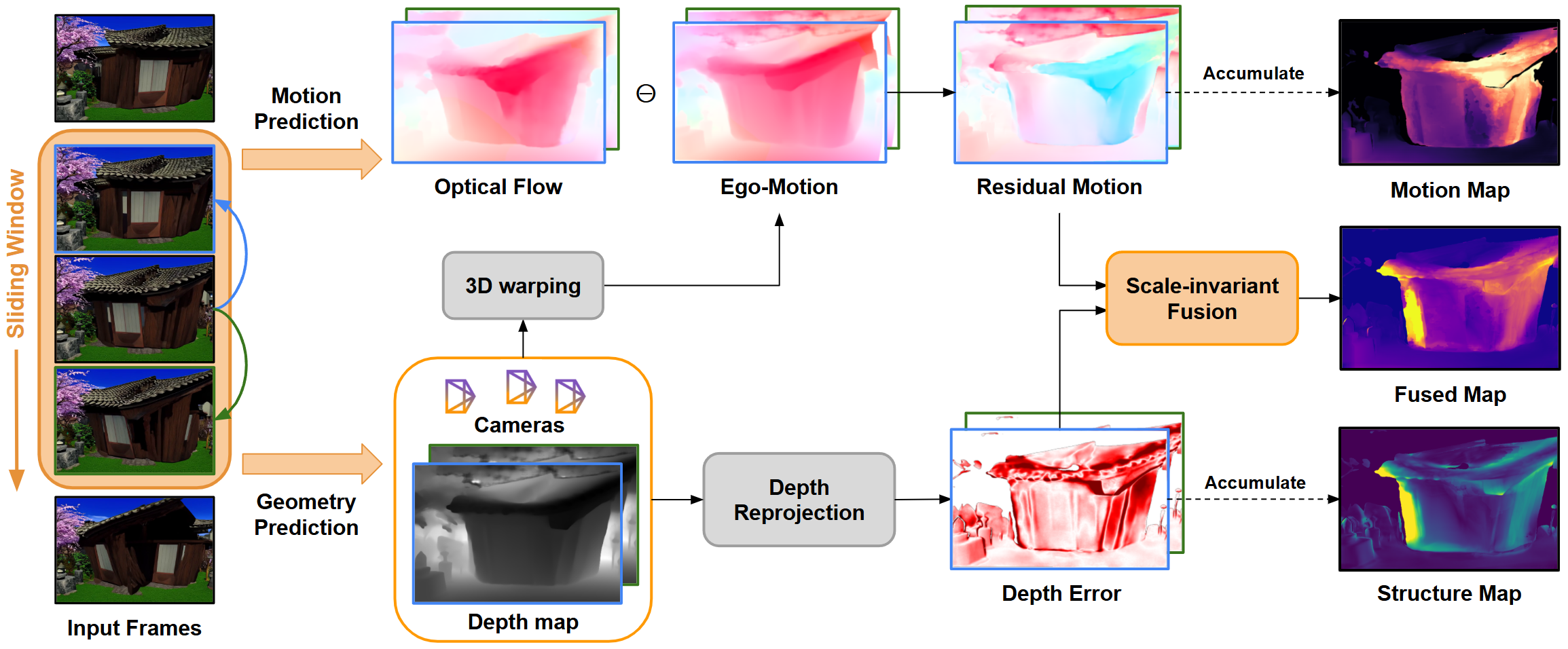}
\caption{
\textbf{\ourmethod{} pipeline.} Within a sliding window, we jointly estimate dense optical flow and 3D geometry (depth and camera pose) for frame pairs. We compute residual motion and depth errors and fuse them into scale-invariant inconsistency maps. Aggregation over the window localizes artifacts in the target frame, while motion and structure maps provide complementary diagnostics.
}
\vspace{-1.5em}
\label{fig:global_pipeline}
\end{figure*}

\section{\ourmethod{}: A Differentiable Geometric Consistency Metric}
\label{methodology}

We introduce \ourmethod{}, a differentiable metric for geometric consistency in generated videos and apply it as guidance to improve 3D consistency for generation.

\subsection{Measuring Geometric Consistency with \ourmethod{}}
\label{sec:metric_definition}
To quantify geometric consistency in static scenes under camera motion, \ourmethod{} uses dense per-pixel correspondences and two complementary cues. Motion consistency measures multi-view rigidity via the residual between optical flow and camera-induced rigid flow in co-visible regions (\ie, geometric deformation). Structure consistency measures object permanence when visibility changes via depth reprojection error, capturing occlusion inconsistency where flow is unreliable. Both cues are then normalized via perspective geometry and fused into a unified, scale-invariant per-pixel inconsistency map.



\cref{fig:global_pipeline} shows the pipeline for metric computation.
\ourmethod{} processes an input video in a sliding window fashion. 
Given $N$ frames in a window, we compute the metric for the center frame $\mathbf{I}_c$. For each pair $(\mathbf{I}_c, \mathbf{I}_i)$ with $i \in \{1,\dots,N\} \setminus \{c\}$, it computes two consistency metrics, motion consistency and structure consistency, and fuses them into a per-pixel error map for the center frame.
This per-pixel map can localize where the deformation occurs in each image.

\myparagraph{Motion consistency.}
Motion consistency measures the residual between optical flow and rigid motion induced by camera in the pixel space.
Given two frames ($\mathbf{I}_c$ and $\mathbf{I}_i$), dense optical flow $\mathbf{F}^{c\rightarrow i}_\text{flow}$ is obtained by an off-the-shelf model~\cite{zhang2025ufm}.
For camera-induced motion, we use a geometry foundation model~\cite{wang2025vggt} to predict depth map $\mathbf{D}_c$, camera intrinsics $\mathbf{K}_c$ (with focal lengths $f_x,f_y$), and relative pose $\mathbf{P}_{c\rightarrow i}{=}(\mathbf{R},\mathbf{T})$.
Then we get residual motion by subtracting the rigid motion from the optical flow:
\vspace{-0.5em}
\begin{gather}
\mathbf{F}_{\text{residual}}(\mathbf{p}) = \mathbf{F}_\text{flow}^{c\rightarrow i}(\mathbf{p})-\mathbf{F}_{\text{rigid}}(\mathbf{p}) \\
\mathbf{F}_{\text{rigid}}(\mathbf{p})
= \pi_{\mathbf{K}_{i}}\!\big(\mathbf{R}\,[\mathbf{D}_c(\mathbf{p})\,\mathbf{K}_c^{-1}\tilde{\mathbf{p}}] + \mathbf{T}\big) - \mathbf{p},
\end{gather}
with $\tilde{\mathbf{p}}$ being the homogeneous coordinate of pixel $\mathbf{p}$ and perspective projection $\pi_{\mathbf{K}}(\cdot)$ with intrinsics $\mathbf{K}$. The residual motion is computed only for co-visible pixels between the center frame and the rest frames, discarding occluded regions where optical flow is not reliable.

\myparagraph{Structure consistency.}
Structure consistency evaluates geometric coherence to handle regions where optical flow is unreliable (\eg occlusions).
Given two frames $\mathbf{I}_c$ and $\mathbf{I}_i$, we reproject the depth map $\mathbf{D}_i$ into the viewpoint of $\mathbf{I}_c$ using the estimated relative pose $\mathbf{P}_{i \rightarrow c}$ and intrinsics $\mathbf{K}_i$.
It yields a warped depth map $\mathbf{D}_{i\rightarrow c}$, utilizing z-buffering to resolve self-occlusions.
The structure consistency at pixel $\mathbf{p}$ is defined as the residual between the predicted and reprojected depth:
\begin{equation}
\Delta z(\mathbf{p}) = \mathbf{D}_c(\mathbf{p}) - \mathbf{D}_{i\rightarrow c}(\mathbf{p}).
\end{equation}

\myparagraph{Scale-invariant fusion.}
Direct fusion of the two signals is non-trivial due to unit mismatch: motion consistency is in pixels, while structure consistency is in depth unit up to scale.
To combine them, we normalize both signals into a dimensionless, scale-invariant space via the pinhole camera model.
Let $(\Delta u, \Delta v) = \mathbf{F}_{\text{residual}}(\mathbf{p})$ be the flow residual and $\Delta z(\mathbf{p})$ be the depth residual.
We define the normalized motion consistency $\textbf{m}(\mathbf{p})$ and structure consistency $\textbf{s}(\mathbf{p})$ as:
\begin{equation}
\textbf{m}(\mathbf{p}) = \sqrt{ \left(\tfrac{\Delta u}{f_x}\right)^2 + \left(\tfrac{\Delta v}{f_y}\right)^2 }, \quad
\textbf{s}(\mathbf{p}) = \left| \tfrac{\Delta z(\mathbf{p})}{\mathbf{D}_c(\mathbf{p})} \right|.
\end{equation}
We fuse these cues into a unified error map $\mathbf{M}_{\text{geo}}$ for the center frame $\mathbf{I}_c$:
\begin{equation}
\mathbf{M}_{\text{geo}}(\mathbf{p}) =
\sqrt{ \big(\mathbbm{1}_{\text{cov}}(\mathbf{p})\cdot \mathbf{m}(\mathbf{p})\big)^2 + \mathbf{s}(\mathbf{p})^2 },
\end{equation}
where $\mathbbm{1}_{\text{cov}}$ is the co-visibility mask.
This allows structure consistency to dominate in occluded regions (where $\mathbbm{1}_{\text{cov}}{=}0$), while utilizing both metrics in co-visible areas to robustly localize deformation.

\myparagraph{Temporal aggregation.}
For each center frame $\mathbf{I}_c$ in a video, we average the pairwise error maps within its sliding window to obtain aggregated frame-level error maps $\{ \mathbf{m}_c, \mathbf{s}_c, \mathbf{M}_{\text{geo},c} \}$.
We then compute the spatial mean of each map to define scalar frame-level scores.
Finally, video-level metrics are obtained by averaging these frame-level scores over all frames in the sequence, yielding three per-video scalars, which we denote by $\mathcal{M}_\text{motion}$, $\mathcal{M}_\text{struct}$, and $\mathcal{M}_\text{geo}$ for motion, structure, and fused geometric consistency, respectively.

\subsection{Training-free Guidance with \ourmethod{}}
\label{sec:method_guidance}

As all components are differentiable, GeCo can serve as a guidance term to improve geometric consistency during video generation, without requiring model fine-tuning. Building on CogVideoX-5B~\cite{yang2024cogvideox} and Frame Guidance~\cite{Jang2025FrameGT}, we update the latent $z_t$ at timestep $t$ by minimizing our \ourmethod{} metric during sampling time:
\begin{equation}
\mathbf{z}_t \; \leftarrow \; \mathbf{z}_t - \eta_t \nabla_{\mathbf{z}_t} \mathcal L_{\text{geo}}\big( \hat{\mathbf{x}}^{\mathcal I}_{0|t} \big),
\label{eq:geo_update}
\end{equation}
where $\eta_t$ is the guidance scale and $\hat{\mathbf{x}}^{\mathcal I}_{0|t}$ represents the approximated clean frames.
The loss $\mathcal{L}_{\text{geo}}$ aggregates the error map $\mathbf{M}_{\text{geo}}$ over a subset of frames $\mathcal I$ and temporal offsets $\mathcal K$:
\begin{equation}
\mathcal{L}_{\text{geo}}(\hat{\mathbf{x}}^{\mathcal I})
= \frac{1}{|\mathcal I||\mathcal K|}
\sum_{c\in\mathcal I} \sum_{k\in\mathcal K}
\sum_{\mathbf{p}\in\Omega} \mathbf{M}_{\text{geo}}^{(c,k)}(\mathbf{p}),
\label{eq:geo_loss}
\end{equation}
where $\mathcal I$ denotes the subset of guidance frames, $\mathcal K$ defines the temporal offsets relative to each center frame $c \in \mathcal I$, and $\Omega$ represents the pixel domain.

As our metric requires clean input images, we estimate the clean latent $\hat{\mathbf{z}}_{0|t}$ from the current noisy state $\mathbf{z}_t$ and predicted velocity $\mathbf{v}_{\theta}$, then decode it via the decoder $D(\cdot)$:
\begin{gather}
\hat{\mathbf{z}}_{0|t} = \sqrt{\bar{\alpha}_t}\, \mathbf{z}_t - \sqrt{1-\bar{\alpha}_t}\, \mathbf{v}_{\theta}(\mathbf{z}_t,t) \\
\hat{\mathbf{x}}^{\mathcal I}_{0|t} = D(\hat{\mathbf{z}}_{0|t})^{\mathcal I},
\label{eq:one_step_vpred}
\end{gather}
Gradients are backpropagated through the frozen decoder and the \ourmethod{} pipeline to update $\mathbf{z}_t$, improving the geometric consistency (\ie, lowering $\mathcal{L}_{\text{geo}}$) throughout the sampling process.
All pretrained backbones used to compute depth~\cite{wang2025vggt} and flow~\cite{zhang2025ufm} are frozen.

To ensure efficiency and stability during sampling, we employ three strategies:
\begin{itemize}
  \item Latent slicing~\cite{Jang2025FrameGT} to decode only a small temporal window $\mathcal I$ for guidance $\mathcal{L}_{\text{geo}}$ computation, instead of using the full sequence.
  \item Recursive denoising~\cite{Lugmayr2022RePaintIU,Bansal2023UniversalGF,Wang2022ZeroShotIR} to repeat updates $R$ times per step to improve convergence
  \item Time-travel~\cite{He2023ManifoldPG} to re-noise the latent after updates, to mitigate accumulated sampling error.
\end{itemize}

\begin{figure*}[tb]
    \centering
    \begin{subfigure}{0.55\linewidth}
        \centering
        \includegraphics[width=\linewidth]{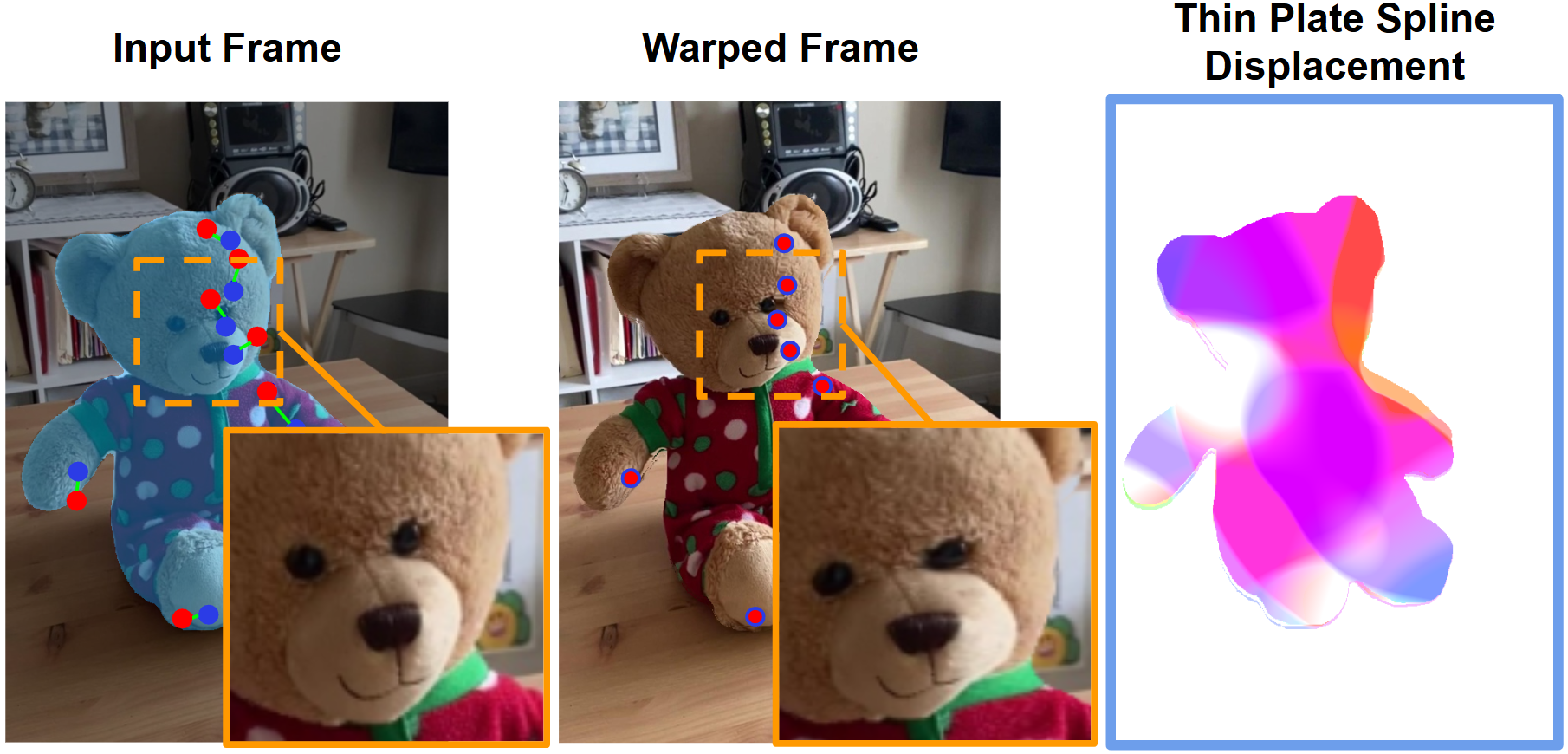}
        \caption{WarpBench deformation process}
        \label{fig:tps}
    \end{subfigure}
    \hfill
    \begin{subfigure}{0.44\linewidth}
        \centering
        \includegraphics[width=\linewidth]{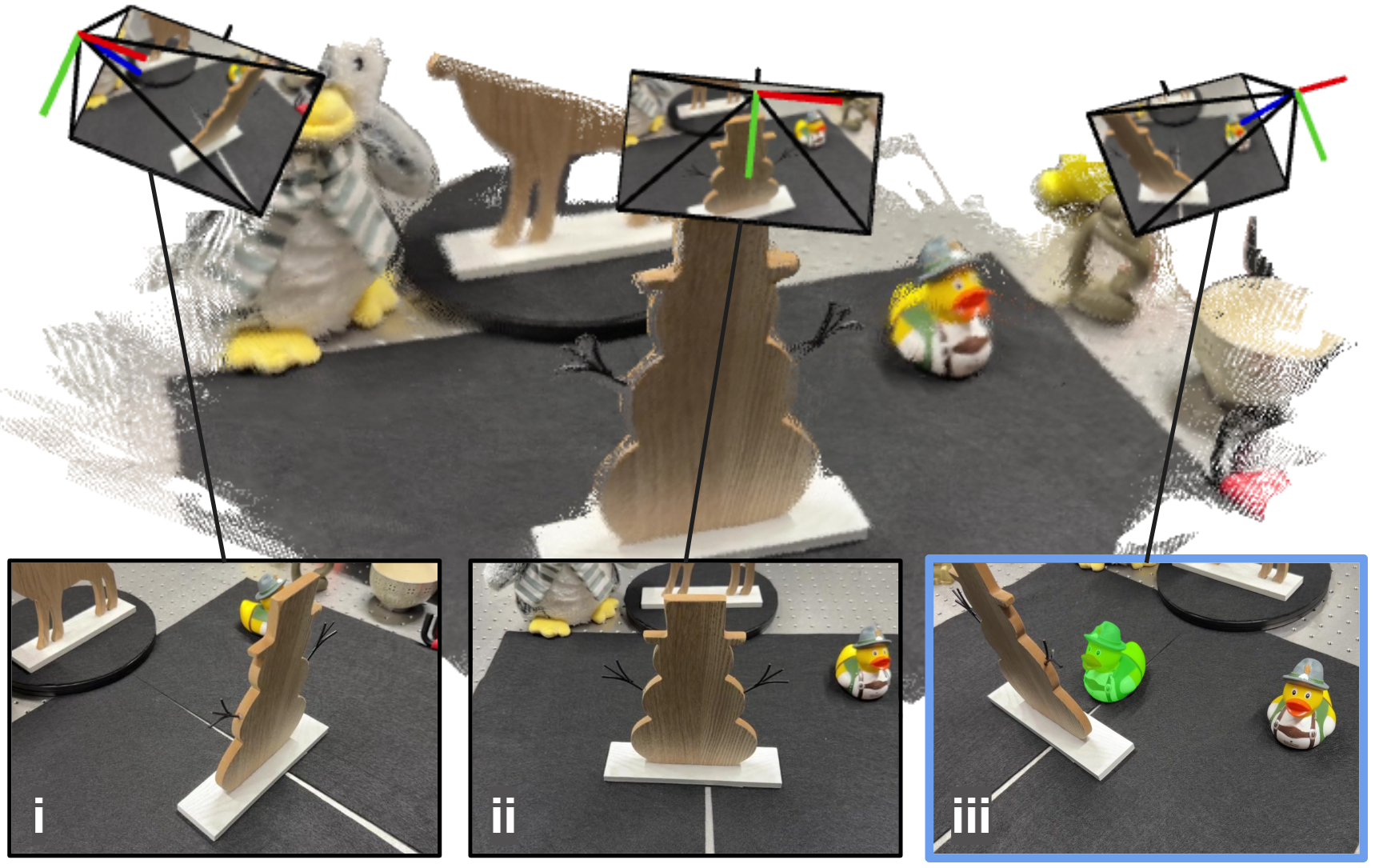}
        \caption{OccluBench sequence}
        \label{fig:occlubench}
    \end{subfigure}
    
    \caption{
        \textbf{Validation datasets.} 
        \textbf{(a) WarpBench}: (Left) Input frame with segmentation mask (cyan), sampled thin-plate spline (TPS) control points (red), and destination points (blue). (Middle) Warped frame after TPS deformation. (Right) Ground-truth displacement field from the deformation. 
        \textbf{(b) OccluBench}: An example sequence where a region in the image center is (i) visible and empty, (ii) occluded, and (iii) re-revealed with a new object, forming a controlled occlusion inconsistency artifact.
    }
    \vspace{-1em}
    \label{fig:combined_benchmarks}
\end{figure*}
\section{\ourmethod{} Validation}
\label{sec:validation}

We validate \ourmethod{} in two steps. First, using controlled datasets to isolate specific artifacts, we demonstrate that its motion and structure cues are complementary and their fusion outperforms baselines (\S\ref{sec:warpbench}--\ref{sec:occlubench}). Second, we confirm the robustness of \ourmethod{} against errors from its underlying off-the-shelf estimators (\S\ref{sec:robustness}). Detailed setups and qualitative visualizations are in the supplementary material.

\subsection{WarpBench for Geometric Deformation}
\label{sec:warpbench}

\myparagraph{WarpBench development.} 
We construct WarpBench using clips from CO3D~\cite{reizenstein21co3d} (\textit{CO3D-Warp}, object-centric) and ScanNet++~\cite{dai2017scannet,yeshwanthliu2023scannetpp} (\textit{ScanNet-Warp}, indoor scenes). To simulate non-rigid deformation artifacts, we inject temporally smooth thin-plate-spline (TPS) warps into foreground regions (\cref{fig:tps}), producing ground-truth dense displacement fields. WarpBench contains 200 clips (4,000 frames). 

\myparagraph{Task1: Frame-level anomaly detection.}
We randomly replace a single frame in a clip with its warped version, and the goal is to identify the anomalous frame. For each clip, we treat each metric as a frame-wise anomaly score and predict the anomalous frame as the one with the largest score in the clip (\ie, $\hat{t} = \arg\max_t \mathcal{M}(t)$). \cref{tab:temporal} shows that motion cue is most reliable on CO3D-Warp and fused cue on ScanNet-Warp. The structure cue is informative for scenes but noisier for object-centric data. MEt3R performs poorly, underscoring the difficulty of detecting subtle deformations with semantic features. WorldScore and TSED improve over MEt3R but trail our cues by a wide margin. WorldScore and TSED improve over MEt3R but trail our cues by a wide margin. WorldScore measures reprojection error, performing well on scene-centric ScanNet-Warp ($81.54\%$) but insensitive to subtle object-level deformation in CO3D-Warp ($32.95\%$). TSED relies on sparse keypoint matching, which lacks dense surface coverage, and performs poorly on both subsets.

\begin{wraptable}{r}{0.56\columnwidth}
\vspace{-3.4em}
\begin{minipage}{0.54\columnwidth}
\centering
\footnotesize
\setlength\tabcolsep{2pt}
\captionof{table}{\textbf{WarpBench: anomaly detection accuracy in \%.} Higher is better. Both of our motion and fused metric shows their competitiveness on both benchmark datasets while MEt3R performs poorly.}
\label{tab:temporal}
\vspace{0.5em}
\scalebox{0.85}{
\begin{tabular}{@{}lcc@{}}
\toprule
Method & CO3D-Warp & ScanNet-Warp \\
\midrule
MEt3R \cite{Asim2025MEt3RMM} & 6.82 & 15.38 \\
WorldScore \cite{duan2025worldscore} & 32.95 & 81.54 \\
TSED \cite{Yu2023PhotoconsistentNVS} & 36.78 & 38.46 \\ \midrule
Structure $\mathcal{M}_\text{struct}$   & 42.05 & 84.62 \\
Motion $\mathcal{M}_\text{motion}$      & \textbf{71.59} & 89.23 \\
Fused $\mathcal{M}_\text{geo}$ & 55.68 & \textbf{92.31} \\
\bottomrule
\end{tabular}
}
\end{minipage}
\vspace{-2em}
\end{wraptable}

\myparagraph{Task 2: Deformation localization.}
Given a real and warped frame pair, each metric outputs an inconsistency map, which we compare against the ground-truth deformation mask. We report AP (\%), IoU (\%), and Spearman’s rank correlation coefficient (SRCC) (See supplementary for details) to measure detection quality, mask overlap, and rank correlation with the true deformation magnitude. As shown in \cref{tab:merged_results} (WarpBench columns), motion is the strongest cue, achieving the highest AP and IoU. MEt3R yields negative SRCC, indicating anti-correlation with deformation magnitude. WorldScore attains positive SRCC and reasonable AP but markedly lower IoU than motion cue ($29.56\%$ vs.\ $44.48\%$ on CO3D-Warp). Fusing depth due does not surpass motion cue alone for pure deformations but remains competitive.

\subsection{OccluBench for Occlusion Inconsistency}
\label{sec:occlubench}

\myparagraph{OccluBench development.}
We construct OccluBench to isolate occlusion inconsistency artifacts: each clip contains a region that is (i) initially visible and empty, (ii) temporarily occluded, and (iii) re-revealed with a new, inconsistent object (\cref{fig:occlubench}). The dataset comprises 30 clips (5,165 frames), with ground-truth artifact masks for 1,654 re-revealed frames annotated using SAM~2~\cite{ravi2024sam2}.

\myparagraph{Task 3: Inconsistent object localization.}
We evaluate artifact localization in re-revealed frames, reporting AP (\%), IoU (\%), and F1 (\%), where AP is threshold-free and implemented as ranking average precision, IoU and F1 are computed at the per-image best threshold, then macro-averaged across $N{=}60$ anchors. Results in \cref{tab:merged_results} (OccluBench columns) confirm the design: motion cue and WorldScore fail almost entirely, as flow cannot track content across occlusion boundaries. In contrast, MEt3R performs far better here than on deformation, while the structure cue effectively localizes artifacts and the fused cue achieves the best overall performance, validating the necessity of fusing both cues.

\begin{table}[tb]
\centering
\footnotesize
\caption{
\textbf{Combined results on WarpBench and OccluBench.} 
We validate our metrics on both benchmarks, WarpBench for deformation and OccluBench for occlusion inconsistency. TSED relies on sparse correspondence without dense prediction, so we report it on the detection task only. Higher is better for all metrics.
}
\vspace{-1em}
\label{tab:merged_results}
\resizebox{\columnwidth}{!}{%
\begin{tabular}{@{}l cccccc ccc@{}}
\toprule
\multirow{3}{*}{Method} & \multicolumn{6}{c}{WarpBench} & \multicolumn{3}{c}{OccluBench} \\
\cmidrule(lr){2-7} \cmidrule(lr){8-10}
& \multicolumn{3}{c}{CO3D-Warp} & \multicolumn{3}{c}{ScanNet-Warp} & \multicolumn{3}{c}{Inconsistency Segmentation} \\
\cmidrule(lr){2-4} \cmidrule(lr){5-7} \cmidrule(lr){8-10}
& AP (\%) $\uparrow$ & IoU (\%) $\uparrow$ & SRCC $\uparrow$ & AP (\%) $\uparrow$ & IoU (\%) $\uparrow$ & SRCC $\uparrow$ & AP (\%) $\uparrow$ & IoU (\%) $\uparrow$ & F1 (\%) $\uparrow$ \\
\midrule
MEt3R \cite{Asim2025MEt3RMM} & 16.26 & 15.95 & -0.176 & 30.34 & 33.13 & -0.351 & 46.51 & 33.97 & 48.57 \\
WorldScore \cite{duan2025worldscore} & 48.70 & 29.56 & 0.253 & 77.51 & 32.45 & 0.332 & 5.36 & 5.77 & 10.61 \\
\midrule
Structure $\mathcal{M}_\text{struct}$  & 25.64 &  3.20 &  0.079 & 51.71 & 14.11 &  0.183 & 62.36 & 51.46 & 66.96 \\
Motion $\mathcal{M}_\text{motion}$      & \textbf{64.90} & \textbf{44.48} & \textbf{0.581} & \textbf{87.12} & \textbf{52.36} & \textbf{0.706} &  4.43 &  5.85 &  8.62 \\
Fused $\mathcal{M}_\text{geo}$  & 60.63 & 41.69 & 0.554 & 82.70 & 48.87 & 0.547 & \textbf{83.48} & \textbf{69.83} & \textbf{81.74} \\
\bottomrule
\end{tabular}%
}
\end{table}

\begin{table}[t]
\centering
\caption{\textbf{Robustness evaluation.} \ourmethod{} consistency scores on artifact-free videos. Fused score divergence between predicted (Pred) and ground-truth (GT) geometry is marginal ($\Delta\!\approx\!0.008$). Crucially, the noise floor on both synthetic and real data is an order of magnitude below generated-video scores (\cref{tab:geoflaw-eval}).}
\vspace{-1em}
\label{tab:robust_exp}
\scalebox{0.8}{%
\footnotesize
\setlength{\tabcolsep}{3pt}
\renewcommand{\arraystretch}{1.02}
\begin{tabular}{lccccccc}
\toprule
\multirow{2}{*}{\textbf{Data}} &
\multicolumn{2}{c}{\textbf{Motion}$\downarrow$} &
\multicolumn{2}{c}{\textbf{Structure}$\downarrow$} &
\multicolumn{2}{c}{\textbf{Fused}$\downarrow$} &
\multirow{2}{*}{\textbf{\makecell{Mean Motion\\(\%/s)}}} \\
\cmidrule(lr){2-3} \cmidrule(lr){4-5} \cmidrule(lr){6-7}
& \textbf{Pred} & \textbf{GT} & \textbf{Pred} & \textbf{GT} & \textbf{Pred} & \textbf{GT} & \\
\midrule
TartanAir V2 Easy & 0.0060 & 0.0009 & 0.0075 & 0.0038 & 0.0066 & 0.0023 &
$32.09\!\pm\!21.76$ \\
TartanAir V2 Hard & 0.0110 & 0.0010 & 0.0088 & 0.0037 & 0.0109 & 0.0024 &
$64.48\!\pm\!38.09$ \\
\midrule
DL3DV & 0.0090 & -- & 0.0172 & -- & 0.0223 & -- &
$121.63\!\pm\!80.56$ \\
\bottomrule
\end{tabular}%
}
\vspace{-1em}
\end{table}

\subsection{\ourmethod{} Robustness Evaluation}
\label{sec:robustness}
As \ourmethod{} builds on off-the-shelf estimators for optical flow, depth, and camera pose, we evaluate the system's robustness by (i) comparing scores computed from predicted versus ground-truth geometry, (ii) measuring the noise floor on real videos, and (iii) stress-testing sensitivity to controlled noise injection.

\myparagraph{Component evaluation.}
We compare \ourmethod{} scores computed from predicted versus ground-truth geometry on TartanAir~v2~\cite{Wang2020TartanAirAD} Easy and Hard splits (\cref{tab:robust_exp}). Evaluating the components directly, the underlying estimators incur 0.757/1.155 flow EPE (px), 0.056/0.061 depth AbsRel, and 0.513/0.748\textdegree\ pose angular error on Easy/Hard, respectively. Despite these degraded predictions on the Hard split, the fused score divergence between predicted and ground-truth remains marginal ($\Delta\!\approx\!0.004$/$0.008$). The predicted fused score (${\sim}0.007$/$0.011$) establishes a practical synthetic noise floor.

\myparagraph{Noise floor on real videos.}
We further run \ourmethod{} on DL3DV~\cite{Ling2023DL3DV10KAL} (\cref{tab:robust_exp}), a dataset of real videos with predominantly rigid camera motion, yielding a predicted fused score of 0.0223. Because real-world data inherently contains more physical noise (e.g., motion blur, complex lighting) than synthetic environments, it is reasonable that this score is slightly higher than the synthetic baseline. Crucially, this real-world noise floor remains an order of magnitude below typical generated-video levels (0.1--0.3 in \cref{tab:geoflaw-eval}). This confirms that elevated scores reflect genuine model-induced geometric failures rather than underlying estimator noise.

\begin{figure*}[tb]
    \centering
    \includegraphics[width=\textwidth]{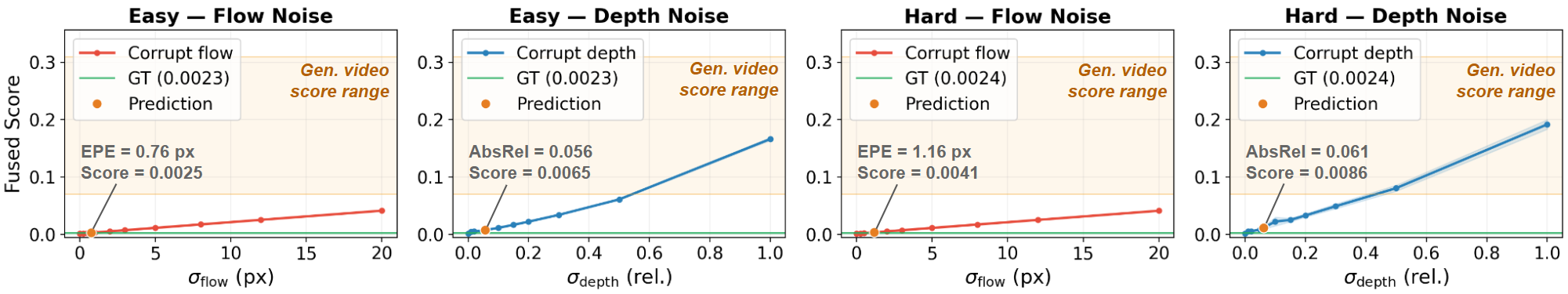}
    \vspace{-1.5em}
\caption{
\textbf{Sensitivity to input noise.} Gaussian noise is added to ground-truth flow ($\sigma_{\text{flow}}$, additive, in pixels) and depth ($\sigma_{\text{depth}}$, multiplicative, in log-space) independently on TartanAir. Orange dots mark the prediction operating point. The score remains near zero well beyond the estimator error level, rising only at substantially larger noise.
}
\vspace{-1.5em}
\label{fig:robustness_gaussian}
\end{figure*}

\myparagraph{Sensitivity to input noise.}
We further stress-test \ourmethod{} by injecting controlled Gaussian noise into ground-truth flow and depth independently on TartanAir~v2 (\cref{fig:robustness_gaussian}). Flow is perturbed additively in pixel space ($\sigma_{\text{flow}}$), and depth multiplicatively in log-space ($\sigma_{\text{depth}}$) so that errors scale with distance. The metric is particularly robust to flow noise: even at $\sigma_{\text{flow}}\!=\!20$~px---over $26\times$/$17\times$ the prediction EPE on Easy/Hard---the fused score remains below the generated-video range. For depth, the score stays below the generated-video range up to $\sigma_{\text{depth}}\!\approx\!0.4$/$0.5$ on Easy/Hard ($7$--$8\times$ the prediction AbsRel). This wide tolerance margin confirms that \ourmethod{} is not brittle to backbone imperfections.

\section{Experiments}
\label{sec:main_exp}
We use \ourmethod{} in two settings: as an evaluation metric to benchmark state-of-the-art video generation models, and as a training-free guidance to reduce geometric artifacts during inference time.
Detailed experimental setups and implementation details are provided in the supplementary material.

\subsection{\ourmethod{}-Eval: Benchmark Suite for T2V Models}
\label{subsec:ourmethod-eval}

\myparagraph{Benchmark design.}
We first evaluate the geometric consistency of text-to-video models by developing \ourmethod{}-Eval benchmark.
To mirror downstream 3D applications, we design four scenarios: \emph{(i)} object-centric scenes, \emph{(ii)} indoor scenes, \emph{(iii)} outdoor scenes, and \emph{(iv)} appearance-complex scenes (\eg, characterized by high-frequency patterns, reflections, and thin structures).
The suite consists of total 80 prompts describing static scenes, split into slow and fast camera trajectories to test both subtle breathing artifacts and multi-view coherence under large viewpoint changes. 

\myparagraph{Models.}
We evaluate recent, state-of-the-art models: 2 commercial models (Sora 2~\cite{brooks2024sora}, Veo 3.1~\cite{veo3techreport}) and 6 open-source models (WAN 2.2~\cite{wan2025wan}, HunyuanVideo~\cite{kong2024hunyuanvideo}, LTX-Video~\cite{HaCohen2024LTXVideo}, and CogVideoX variants~\cite{hong2022cogvideo,yang2024cogvideox}). 
We generate 4 videos per each prompt across 8 models, yielding 2,560 videos in total.

\newcommand{\bestc}[1]{\cellcolor{green!35}\textbf{#1}}
\newcommand{\besto}[1]{\cellcolor{green!20}\textbf{#1}}
\newcommand{\highmotion}[1]{\cellcolor{red!20}#1}

\begin{table}[t]
\centering
\scriptsize
\setlength{\tabcolsep}{1.5pt}
\renewcommand{\arraystretch}{0.95}
\caption{\textbf{\ourmethod-Eval across scenarios.}
Per-model motion inconsistency (Mot.), structure inconsistency (Str.), and fused \ourmethod{} scores (lower is better) together with Total Motion (TM, \%) and Mean Motion (MM, \%/s; mean$\pm$std over clips). Commercial models (top block) and open-source models (bottom block) are compared across four scenarios. Darker green cells indicate the best scores among commercial models, lighter green cells indicate the best scores among open-source models, and red cells highlight models with very low TM and MM, where a high consistency score may be partially achieved by under-shooting motion (near-static generations).}
\vspace{-1em}
\label{tab:geoflaw-eval}
\begin{minipage}{0.49\columnwidth}
\centering
{\scriptsize (a) Object-centric} \par\vspace{2pt}
\resizebox{\linewidth}{!}{%
\begin{tabular}{@{}l ccc cc@{}}
\toprule
Model & Mot. $\downarrow$ & Str. $\downarrow$ & Fused $\downarrow$ & TM (\%) & MM (\%/s) \\
\midrule
Sora 2            & 0.016 & \bestc{0.063} & \bestc{0.069} & $24.97 \pm 42.68$ & $6.29 \pm 10.76$ \\
Veo 3.1           & \bestc{0.012} & 0.087 & 0.090 & $52.48 \pm 34.84$ & $6.59 \pm 4.38$  \\
\midrule
WAN 2.2           & 0.027 & 0.131 & 0.140 & $87.11 \pm 91.10$ & $17.42 \pm 18.22$ \\
HunyuanVideo      & 0.014 & 0.333 & 0.336 & $34.29 \pm 35.06$ & $6.43 \pm 6.57$ \\
CogVideoX1.5      & 0.008 & 0.054 & 0.056 & \highmotion{$9.16 \pm 11.55$}  & \highmotion{$1.83 \pm 2.31$}  \\
CogVideoX-5B      & 0.027 & 0.073 & 0.080 & $59.81 \pm 77.73$ & $19.94 \pm 25.91$ \\
CogVideoX-2B      & 0.018 & 0.053 & 0.059 & $30.57 \pm 34.77$ & $10.19 \pm 11.59$ \\
LTX-Video         & \besto{0.005} & \besto{0.024} & \besto{0.026} & \highmotion{$13.01 \pm 20.07$} & \highmotion{$3.25 \pm 5.02$} \\
\bottomrule
\end{tabular}%
}
\end{minipage}
\hfill
\begin{minipage}{0.49\columnwidth}
\centering
{\scriptsize (b) Indoor} \par\vspace{2pt}
\resizebox{\linewidth}{!}{%
\begin{tabular}{@{}l ccc cc@{}}
\toprule
Model & Mot. $\downarrow$ & Str. $\downarrow$ & Fused $\downarrow$ & TM (\%) & MM (\%/s) \\
\midrule
Sora 2            & \bestc{0.026} & \bestc{0.162} & \bestc{0.169} & $71.12 \pm 123.63$ & $17.93 \pm 31.17$ \\
Veo 3.1           & 0.030 & 0.223 & 0.233 & $136.06 \pm 113.56$ & $17.10 \pm 14.27$ \\
\midrule
WAN 2.2           & 0.039 & 0.269 & 0.279 & $106.79 \pm 111.28$ & $21.36 \pm 22.26$ \\
HunyuanVideo      & 0.027 & 0.274 & 0.281 & $122.06 \pm 139.96$ & $22.89 \pm 26.24$ \\
CogVideoX1.5      & 0.010 & 0.115 & 0.117 & \highmotion{$14.22 \pm 14.12$} & \highmotion{$2.84 \pm 2.82$} \\
CogVideoX-5B      & 0.025 & 0.128 & 0.136 & $57.14 \pm 68.53$ & $19.05 \pm 22.84$ \\
CogVideoX-2B      & 0.022 & 0.117 & 0.124 & $47.14 \pm 43.65$ & $15.71 \pm 14.55$ \\
LTX-Video         & \besto{0.009} & \besto{0.070} & \besto{0.073} & \highmotion{$21.27 \pm 33.21$} & \highmotion{$5.32 \pm 8.30$} \\
\bottomrule
\end{tabular}%
}
\end{minipage}

\vspace{2pt}

\begin{minipage}{0.49\columnwidth}
\centering
{\scriptsize (c) Outdoor} \par\vspace{2pt}
\resizebox{\linewidth}{!}{%
\begin{tabular}{@{}l ccc cc@{}}
\toprule
Model & Mot. $\downarrow$ & Str. $\downarrow$ & Fused $\downarrow$ & TM (\%) & MM (\%/s) \\
\midrule
Sora 2            & 0.032 & 0.258 & 0.265 & $63.62 \pm 97.34$ & $16.04 \pm 24.54$ \\
Veo 3.1           & \bestc{0.026} & \bestc{0.209} & \bestc{0.217} & $161.61 \pm 137.62$ & $20.31 \pm 17.29$ \\
\midrule
WAN 2.2           & 0.054 & 0.423 & 0.437 & $159.98 \pm 179.82$ & $32.00 \pm 35.96$ \\
HunyuanVideo      & 0.028 & 0.232 & 0.240 & $118.64 \pm 150.64$ & $22.25 \pm 28.24$ \\
CogVideoX1.5      & \besto{0.009} & \besto{0.043} & \besto{0.047} & \highmotion{$24.44 \pm 35.76$} & \highmotion{$4.89 \pm 7.15$} \\
CogVideoX-5B      & 0.032 & 0.172 & 0.180 & $81.52 \pm 86.97$ & $27.17 \pm 28.99$ \\
CogVideoX-2B      & 0.015 & 0.063 & 0.068 & $44.15 \pm 39.60$ & $14.72 \pm 13.20$ \\
LTX-Video         & 0.011 & 0.068 & 0.071 & \highmotion{$27.32 \pm 37.89$} & \highmotion{$6.83 \pm 9.47$} \\
\bottomrule
\end{tabular}%
}
\end{minipage}
\hfill
\begin{minipage}{0.49\columnwidth}
\centering
{\scriptsize (d) Appearance-complex} \par\vspace{2pt}
\resizebox{\linewidth}{!}{%
\begin{tabular}{@{}l ccc cc@{}}
\toprule
Model & Mot. $\downarrow$ & Str. $\downarrow$ & Fused $\downarrow$ & TM (\%) & MM (\%/s) \\
\midrule
Sora 2            & 0.035 & \bestc{0.195} & \bestc{0.206} & $56.58 \pm 90.84$ & $14.26 \pm 22.90$ \\
Veo 3.1           & \bestc{0.030} & 0.200 & 0.210 & $90.61 \pm 80.47$ & $11.39 \pm 10.11$ \\
\midrule
WAN 2.2           & 0.047 & 0.193 & 0.212 & $104.36 \pm 123.70$ & $20.87 \pm 24.74$ \\
HunyuanVideo      & 0.031 & 0.180 & 0.193 & $53.65 \pm 58.44$ & $10.06 \pm 10.96$ \\
CogVideoX1.5      & 0.012 & 0.176 & 0.181 & \highmotion{$11.28 \pm 10.04$} & \highmotion{$2.26 \pm 2.01$} \\
CogVideoX-5B      & 0.041 & 0.110 & 0.126 & $53.35 \pm 54.24$ & $17.78 \pm 18.08$ \\
CogVideoX-2B      & 0.024 & 0.080 & 0.091 & $39.47 \pm 38.52$ & $13.16 \pm 12.84$ \\
LTX-Video         & \besto{0.007} & \besto{0.039} & \besto{0.041} & \highmotion{$13.51 \pm 29.38$} & \highmotion{$3.38 \pm 7.34$} \\
\bottomrule
\end{tabular}%
}
\end{minipage}
\vspace{-1.5em}
\end{table}

\myparagraph{Evaluation protocol.}
To ensure fair comparison across different models with their own native generation settings, we compute \ourmethod{} scores on overlapping $\approx 3$-second windows resampled to a maximum of $f_{\text{eval}}{=}8$ FPS.
Within each window, we compute motion, structure, and fused scores averaged over pixels with valid depth and covisibility.
Final clip-level scores are derived from the frame-weighted average of these windows, and we report the mean across all 320 clips (80 prompts $\times$ 4 seeds) per model.

\myparagraph{Motion magnitude.}
High consistency scores can stem from valid geometry or simply a lack of motion in generated videos. Therefore, geometric consistency metrics should be jointly analyzed with motion to distinguish truly consistent generation from degenerate, static outputs.
We first quantify motion using the per-frame optical flow magnitude normalized by the image diagonal ($m_t$), and introduce two metrics: Total Motion is the cumulative flow ($\sum m_t$) over the clip, representing the overall dynamism and amount of visual change independent of FPS.
Mean Motion divides total motion by duration ($\text{Total Motion}/S$), representing the average rate of visual change (\ie, apparent camera speed).

\begin{figure*}[tb]
    \centering
    \includegraphics[width=\textwidth]{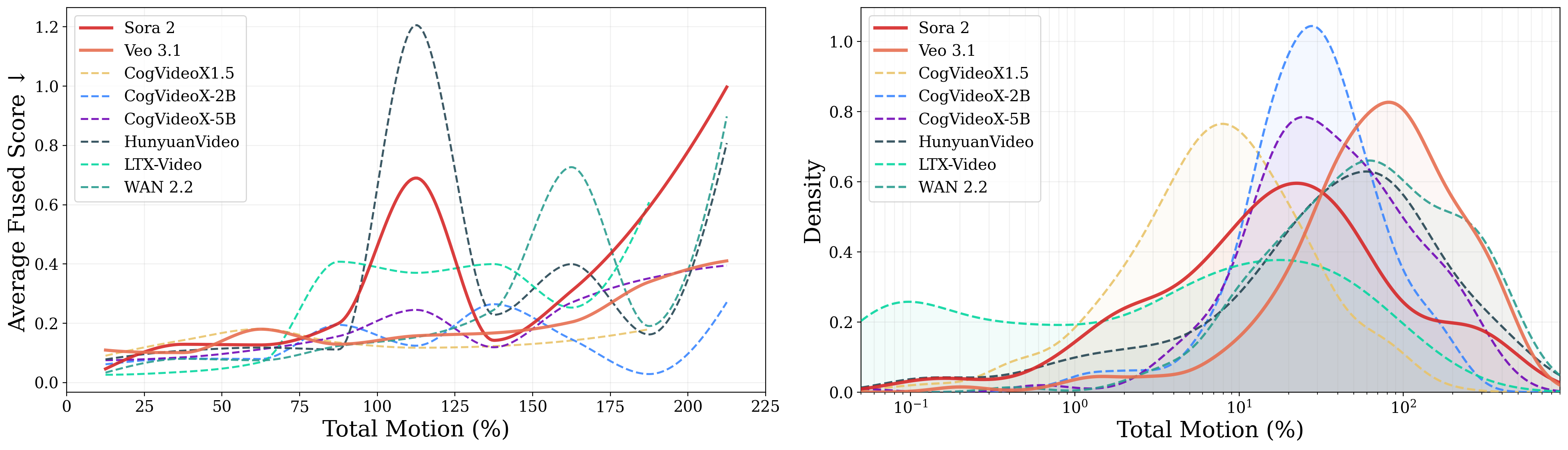}
    \vspace{-1.5em}
    \caption{\textbf{Motion-dependent consistency and dynamism distribution.} \textbf{Left:} Average \ourmethod{} Fused Score versus Total Motion. At lower motion levels (<100\%), most models maintain stable scores, but in higher motion levels (>100\%), several models struggle with obvious artifacts, leading to sharply degraded consistency scores. \textbf{Right:} Motion distribution of generated videos. Most models concentrate motion between 10\%--100\%, while Veo 3.1 notably tends to generate videos with higher dynamism.}
    \vspace{-2em}
    \label{fig:motion-score}
\end{figure*}

\begin{wrapfigure}{r}{0.5\textwidth}
  \centering
  \vspace{-2.3em}
  \includegraphics[width=\linewidth]{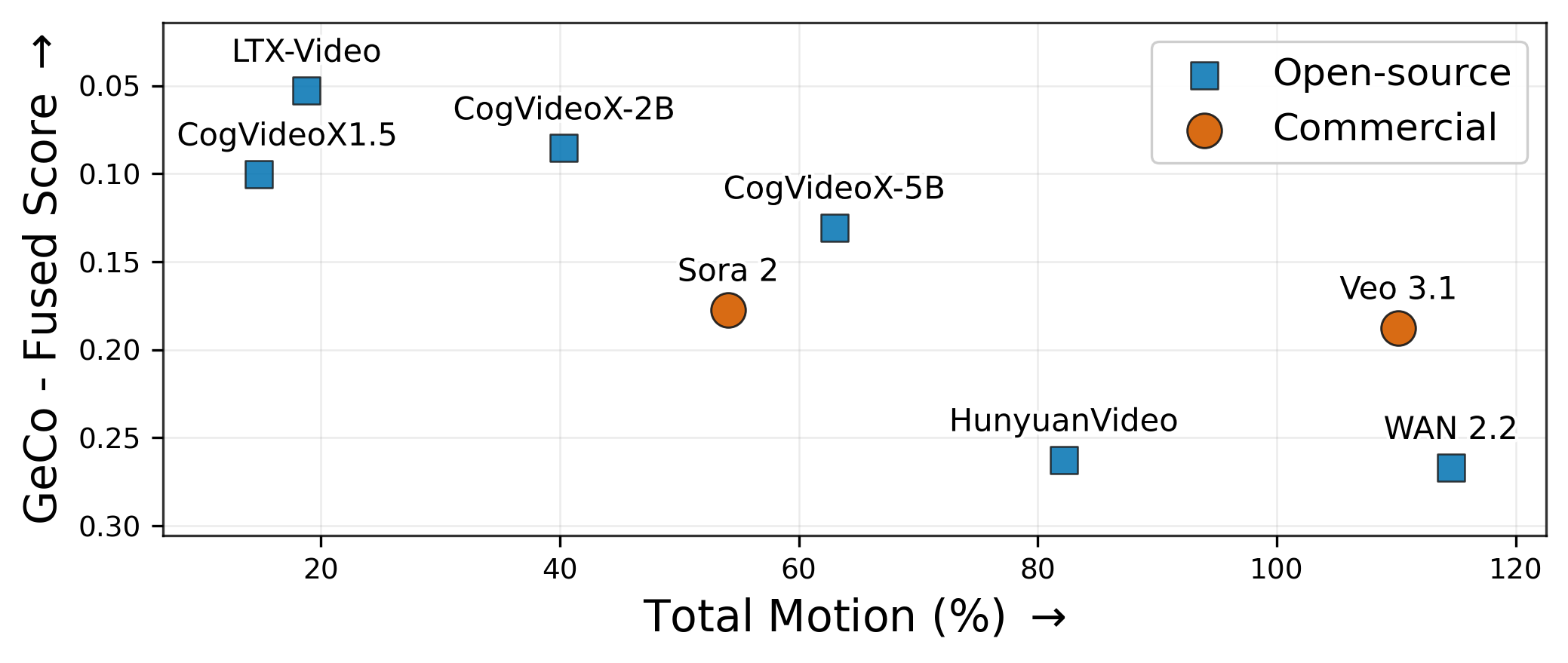}
  \vspace{-2em}
  \caption{\textbf{Motion--consistency Pareto analysis.} Fused \ourmethod{} score versus Total Motion, averaged across all four evaluation scenarios from \cref{tab:geoflaw-eval}. The inverted y-axis places generations with high dynamism and geometric consistency in the top-right quadrant.}
  \label{fig:pareto_front}
  \vspace{-2.5em}
\end{wrapfigure}

\myparagraph{Motion--consistency analysis.}
\cref{tab:geoflaw-eval} details the quantitative benchmark scores across scenarios. As some open-source models achieve better consistency scores simply by generating near-static videos, we contextualize these results using a motion--consistency Pareto front (\cref{fig:pareto_front}). Among large-scale models, Veo~3.1 and Sora~2 achieve similarly strong consistency, though Veo sustains notably higher dynamism. Conversely, models like LTX-Video anchor the lower-error end of the frontier primarily by operating in a highly conservative, low-motion regime. Notably, CogVideoX-5B acts as a strong middle ground for open-weight models, achieving surprisingly good consistency while maintaining an intermediate dynamic range. Furthermore, finer-level analysis (\cref{fig:motion-score}) reveals that while Veo leans heavily toward high dynamism and LTX-Video toward conservatism, most models suffer from severely degraded consistency scores when video dynamics increase (TM$>$100\%). Against this trend, Veo~3.1 and CogVideoX1.5 prove to be the most stable under highly dynamic conditions.

\myparagraph{Real vs.\ generated videos.}
We compare different metric score averages on 30 real videos from DL3DV and 30 generated videos (5 per model from \ourmethod{}-Eval), with comparable motion magnitudes (Total Motion, \%: real $47.88\pm35.05$ vs.\ generated $52.44\pm46.68$). We use Cohen's~$d$ to measure the standardized gap between real and generated scores, where positive values indicate real videos score as more consistent. As shown in Tab.~\ref{tab:real_vs_gen}, \ourmethod{}, WorldScore, and TSED all yield positive $d$, whereas MEt3R is negative ($d{=}-0.361$). We do not expect a metric to separate real from generated videos, since generators can produce geometrically consistent results; rather, a valid metric should not systematically rate generated videos as more consistent than real ones. MEt3R's negative $d$ likely stems from its focus on semantic feature consistency, which is less sensitive to the geometric artifacts.

\begin{table}[t]
\centering
\caption{\textbf{Real vs.\ generated score gap.} Average scores on 30 real (DL3DV) and 30 generated videos; Cohen's~$d$ measures the standardized gap, positive when real videos score as more consistent. $\downarrow$/$\uparrow$: lower/higher score is more consistent.}
\vspace{-0.5em}
\label{tab:real_vs_gen}
\scalebox{0.8}{%
\footnotesize
\setlength{\tabcolsep}{3pt}
\renewcommand{\arraystretch}{1.02}
\begin{tabular}{@{}lccc@{}}
\toprule
Metric & Real ($n{=}30$) & Generated ($n{=}30$) & Cohen's $d$ \\
\midrule
MEt3R\,$\downarrow$       & $0.9329 \pm 0.0308$ & $0.9119 \pm 0.0763$ & $-0.361$ \\
WorldScore\,$\downarrow$  & $0.4592 \pm 0.2572$ & $0.6357 \pm 0.6565$ & $+0.354$ \\
TSED\,$\uparrow$          & $1.0000 \pm 0.0000$ & $0.9870 \pm 0.0573$ & $+0.322$ \\
\midrule
\ourmethod{}\,$\downarrow$ & $0.0999 \pm 0.1887$ & $0.1809 \pm 0.2682$ & $+0.349$ \\
\bottomrule
\end{tabular}%
}
\vspace{-0.5em}
\end{table}

\subsection{\ourmethod{}‑Guided Inference}
\label{subsec:\ourmethod{}-guidance}

We apply \ourmethod{} as training-free guidance to CogVideoX-5B (\cref{sec:method_guidance}). Using 45 benchmark prompts, we generate paired videos with and without guidance under identical settings (details in the supplementary) and evaluate them using our metrics alongside MEt3R~\cite{Asim2025MEt3RMM} to quantify the change in multiple metrics.

\begin{figure*}[t]
\centering
    \includegraphics[width=0.99\linewidth]{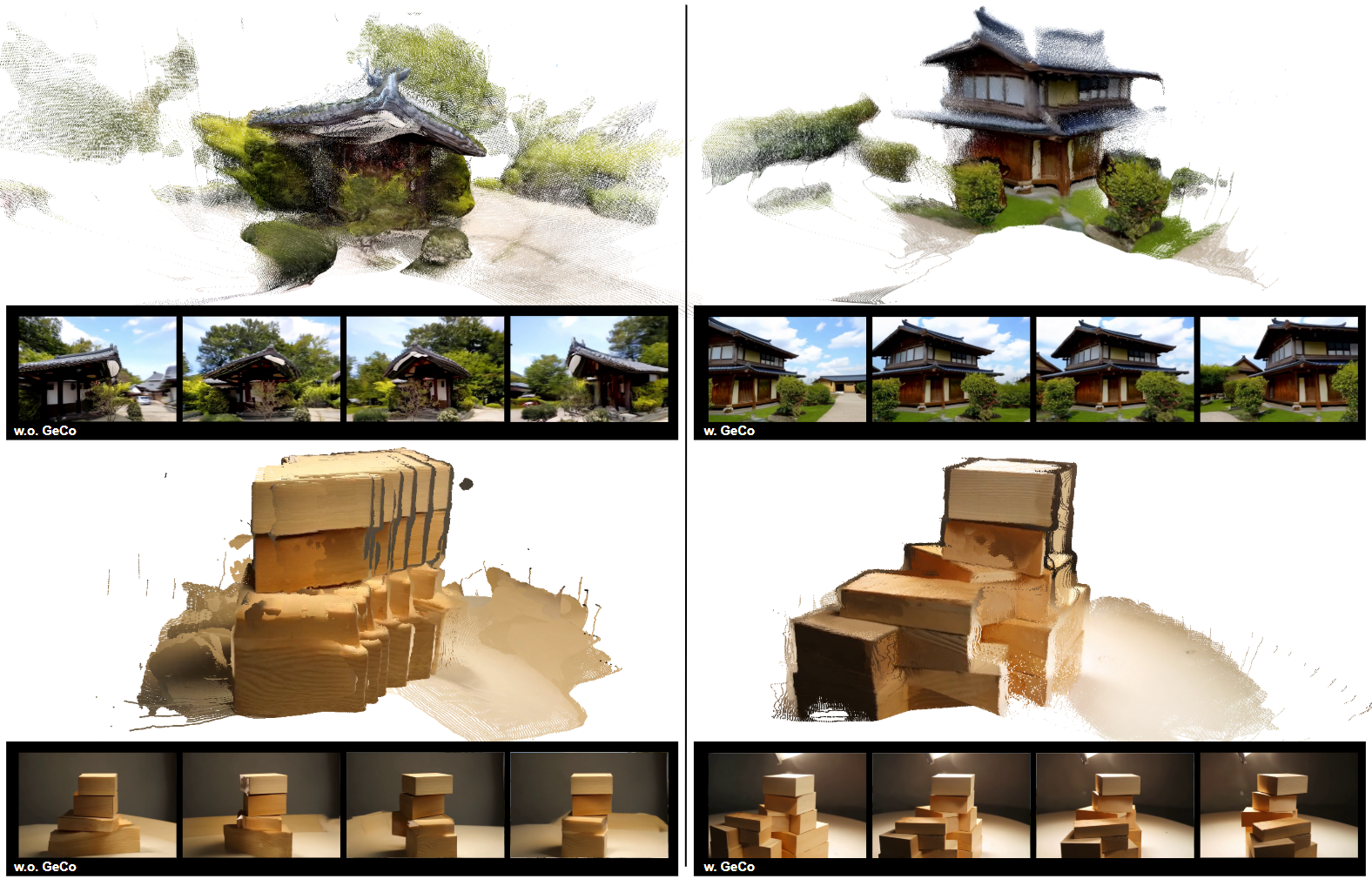}
\vspace{-0.5em}
\caption{
\textbf{\ourmethod{} guidance improves geometric consistency for 3D reconstruction.}
Top: 3D reconstructions from videos generated by CogVideoX‑5B without (left) and with \ourmethod{} guidance (right). 
Bottom: corresponding video frames.
Both guided videos yield cleaner geometry with fewer deformation and drifting artifacts across views, which enables higher reconstruction quality.
}\vspace{-0.5em}
 \label{fig:guidance}
\end{figure*}

As reported in \cref{tab:guidance_exp}, \ourmethod{} guidance consistently improves all scores by reducing both structure and motion errors. Notably, guided videos exhibit slightly higher motion, indicating that \ourmethod{} enforces geometric consistency without sacrificing dynamism. Finally, we perform 3D reconstruction on the generated videos using VGGT. Qualitative results (\cref{fig:guidance}) demonstrate that guided videos yield significantly cleaner reconstructions, featuring reduced drifting and fewer artifacts around thin structures.

\begin{table}[tb]
\centering
\footnotesize
\setlength{\tabcolsep}{4pt}
\renewcommand{\arraystretch}{1.0} 
\caption{
\textbf{\ourmethod{}‑guided sampling improves geometric consistency.}
We compare CogVideoX‑5B with and without \ourmethod{} guidance on a set of 45 prompts.
Lower is better for consistency metrics. Guided videos exhibit slightly higher motion (TM, MM), confirming that \ourmethod{} improves consistency without hurting dynamism.
}
\vspace{-0.5em}
\label{tab:guidance_exp}
\resizebox{0.85\columnwidth}{!}{
\begin{tabular}{@{}l cccc cc@{}}
\toprule
Method & MEt3R$\downarrow$ & Structure$~\downarrow$ & Motion$~\downarrow$ & Fused$~\downarrow$ & TM (\%) & MM (\%/s) \\
\midrule
Without guidance  & 0.100 & 0.136 & 0.032 & 0.147 & $74.35{\pm}68.09$ & $24.78{\pm}22.70$ \\
\textbf{With guidance}  & \textbf{0.098} & \textbf{0.126} & \textbf{0.029} & \textbf{0.135} & $\mathbf{83.16{\pm}98.36}$ & $\mathbf{27.72{\pm}32.79}$ \\
\bottomrule
\end{tabular}%
}
\vspace{-1em}
\end{table}

\begin{figure*}[t]
    \centering
    \includegraphics[width=\linewidth]{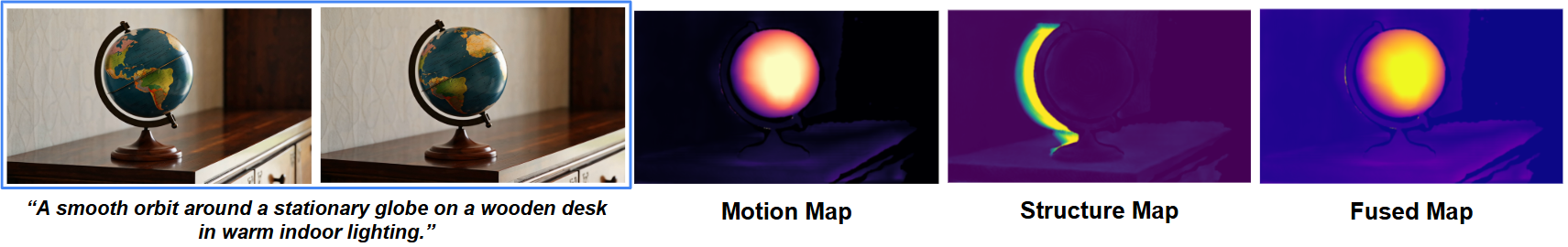}
    \vspace{-2em}
\caption{\textbf{``The globe that can’t be stopped.''}
A common failure mode that models consistently make the globe rotate despite static prompts.
\ourmethod{} localizes this spurious object motion on the globe surface, clearly separating it from the intended egocentric camera motion.}
    \label{fig:globe}
    \vspace{-2em}
\end{figure*}

\subsection{Findings on Video Generation Models}
\label{subsec:t2vfinding}

\myparagraph{The globe that can’t be stopped.}
We identify a persistent failure mode where video models struggle to generate static objects under camera motion, likely due to training data bias (\eg, spinning globes). As shown in \cref{fig:globe}, models consistently animate the globe despite static prompts, while \ourmethod{} correctly identifies this spurious object motion as a geometric inconsistency. We believe a possible reason for the failure mode is the bias in training data, where most examples of globes appear in motion, leading models to conflate object persistence with deformation or drift. 

\begin{figure*}[t]
    \centering
    \includegraphics[width=\linewidth]{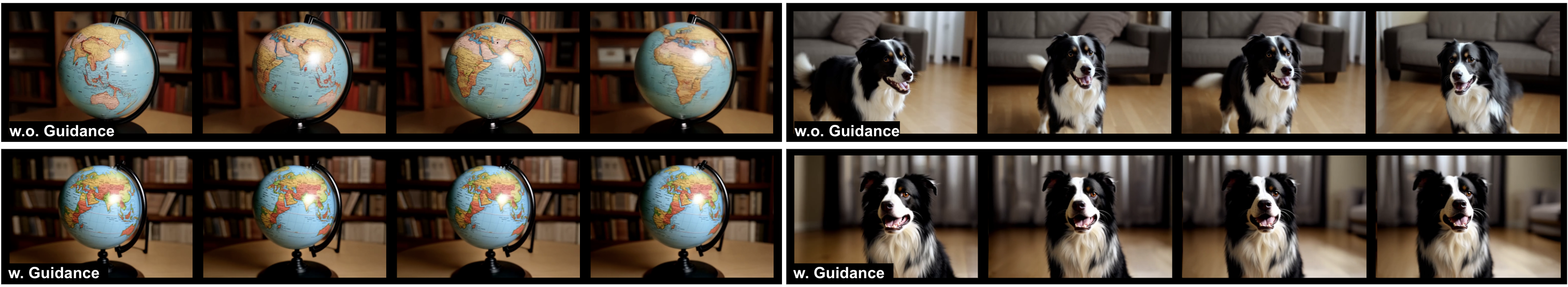}
    \vspace{-2em}
\caption{\textbf{Stopping the globe and freezing the dog with \ourmethod{}-guidance.} We compare video generations for prompts specifying a camera orbiting a nominally static globe (left) and a static dog (right). Without guidance, the model introduces spurious object motion, causing the globe to spin and the dog to move. Applying \ourmethod{} guidance effectively suppresses this non-ego motion, enforcing geometric consistency where the subject remains rigid while the camera orbits, producing a ``bullet-time'' effect.}
    \label{fig:failure_mode}
    \vspace{-1em}
\end{figure*}

\myparagraph{Freezing the spurious motion.} To empirically evaluate our assumption, we test prompts that should be sparse in video training data such as "a camera orbiting a dog that has no motion." Models typically fail to keep the subject rigid.
However, by using \ourmethod{} as a guidance term, we can strongly suppress this non-ego motion.
As illustrated in \cref{fig:failure_mode}, our guidance effectively freezes the subject, creating a bullet-time effect where the camera orbits while the object remains rigid. 

\section{Discussion}
\myparagraph{Limitations.}
\ourmethod{}'s assumption of a mostly rigid scene means dynamic contents can trigger false positives by penalizing true object motion. Addressing this by applying motion masks to static backgrounds is a promising direction for future work. Beyond scene constraints, deploying \ourmethod{} for training-free guidance increases generation time (e.g., from 2 to 19 minutes on an H200 GPU). Because this overhead stems from iterative sampling and backpropagation rather than the metric itself, integrating efficient samplers or lightweight estimators could substantially reduce it. Finally, our metric inherits the limitations of its off-the-shelf estimators. Although we observed mild prediction noise, it remains well below typical artifact severity, meaning \ourmethod{} will seamlessly yield more accurate localization as geometry foundation models advance.

\myparagraph{Conclusion.}
We present \ourmethod{}, an interpretable metric that fuses optical flow and depth priors to localize geometric artifacts in generated videos. After validating robustness on controlled datasets, we use it to benchmark state-of-the-art models at scale and identify persistent failure modes. We further show \ourmethod{} can act as a training-free guidance loss, improving geometric consistency, reducing spurious motion, and benefiting downstream 3D reconstruction. As \ourmethod{} is fully differentiable, it also supports finetuning and distillation. Overall, \ourmethod{} provides actionable diagnostics for geometric consistency and supports applications such as 3D asset creation and world simulation.

\section*{Acknowledgements}
This project is supported in part by NIH grant R01HD104969 and NSF grant CRCNS-2309041. The views and conclusions contained herein are those of the authors and do not represent the official policies or endorsements of these institutions.

\clearpage
\setcounter{page}{1}
\setcounter{figure}{0}
\renewcommand{\thefigure}{\Alph{figure}}
\setcounter{table}{0}
\renewcommand{\thetable}{\Alph{table}}
\appendix

\begin{center}
    {\Large \bfseries \ourmethod{}: Evaluating Geometric Consistency \\ [0.2em] 
    for Video Generation via Motion and Structure \par}
    \vspace{1em}
    {\large Supplementary Material \par}   
    \vspace{-2em}
\end{center}

\begin{figure}[!h]
\centering
\includegraphics[width=0.99\linewidth]{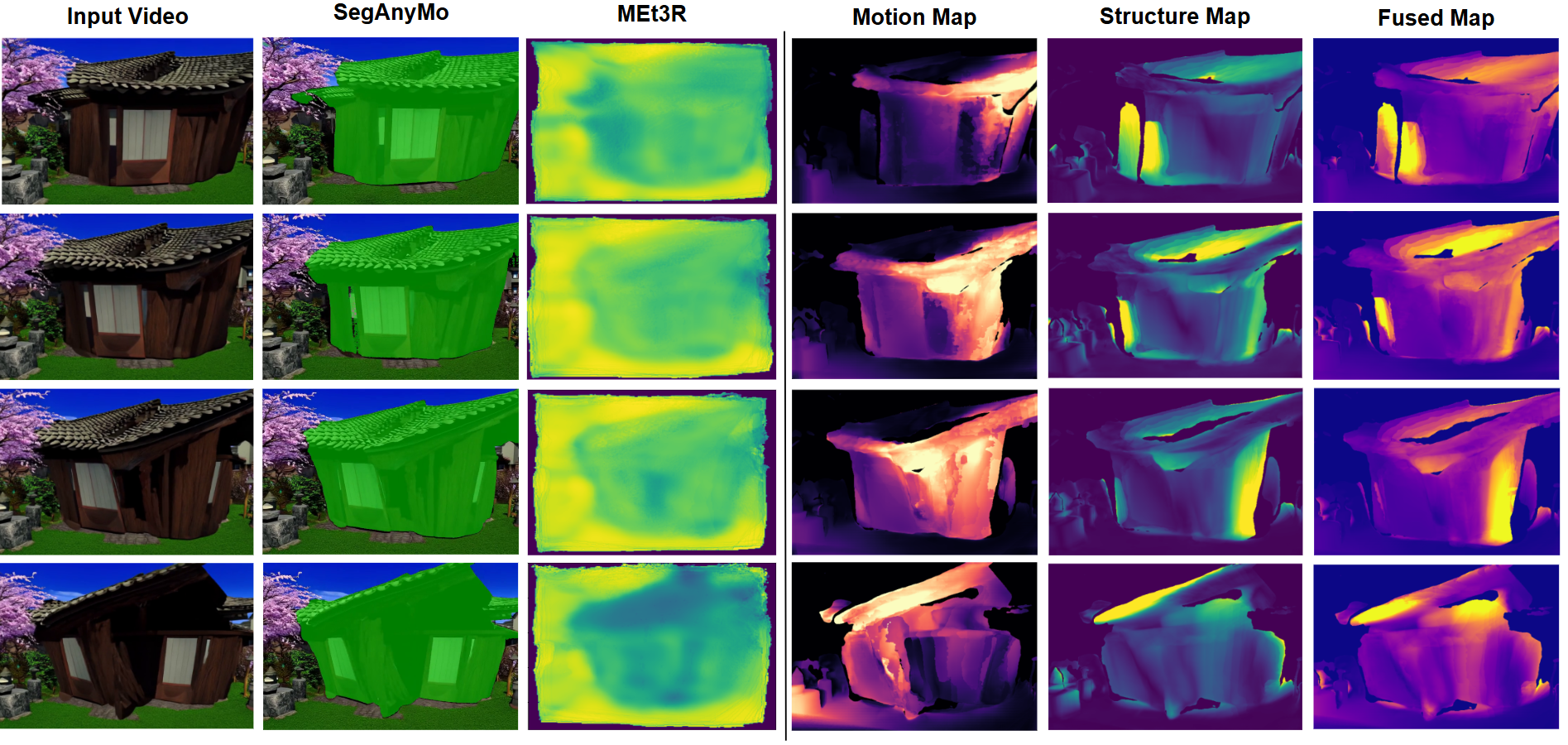}
\caption{\textbf{Qualitative comparison on deformation artifacts.} SegAnyMo~\cite{huang2025segment} fails to predict localized motion, while MEt3R~\cite{Asim2025MEt3RMM} produces a blurred score map without region-level detail. In contrast, our method, \ourmethod{}, produces interpretable inconsistency maps that precisely highlight subtle, localized geometric distortions rather than masking entire objects.}
\vspace{-1em}
\label{fig:interpretability_comparison}
\end{figure}

\begin{figure}[tb]
\centering
    \includegraphics[width=\linewidth]{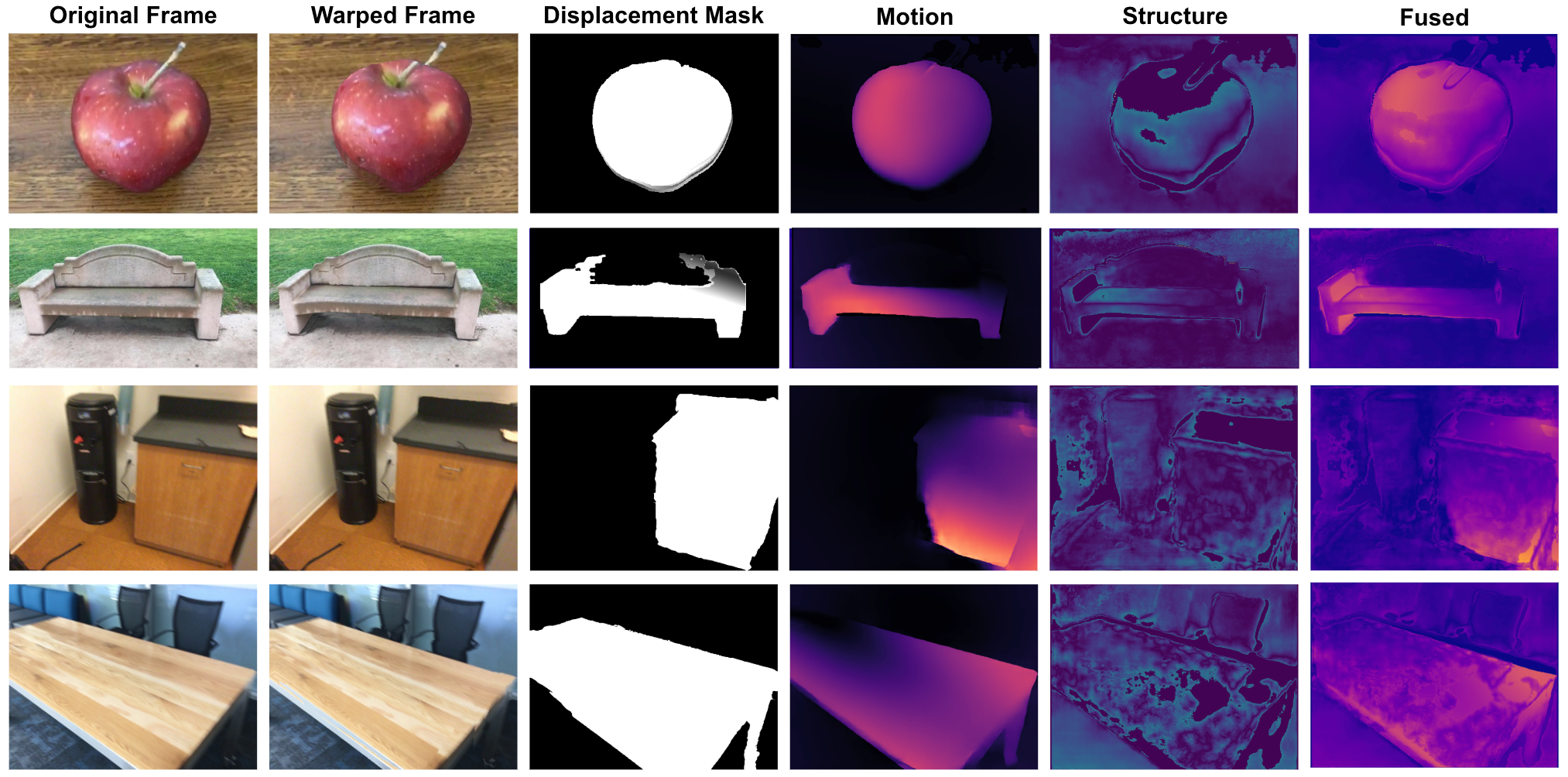}
\caption{
\textbf{Motion cue validation on WarpBench.}
Top two rows: examples from CO3D-Warp; Bottom two rows: examples from ScanNet-Warp. In all examples, the motion map accurately highlights the deformation region (aligning with the displacement mask). In contrast, the structure cue remains insensitive to these surface-level deformations, as they induce only negligible depth variations.
}
\label{fig:warp_vis}
\end{figure}

\begin{figure}[tb]
\centering
    \includegraphics[width=\linewidth]{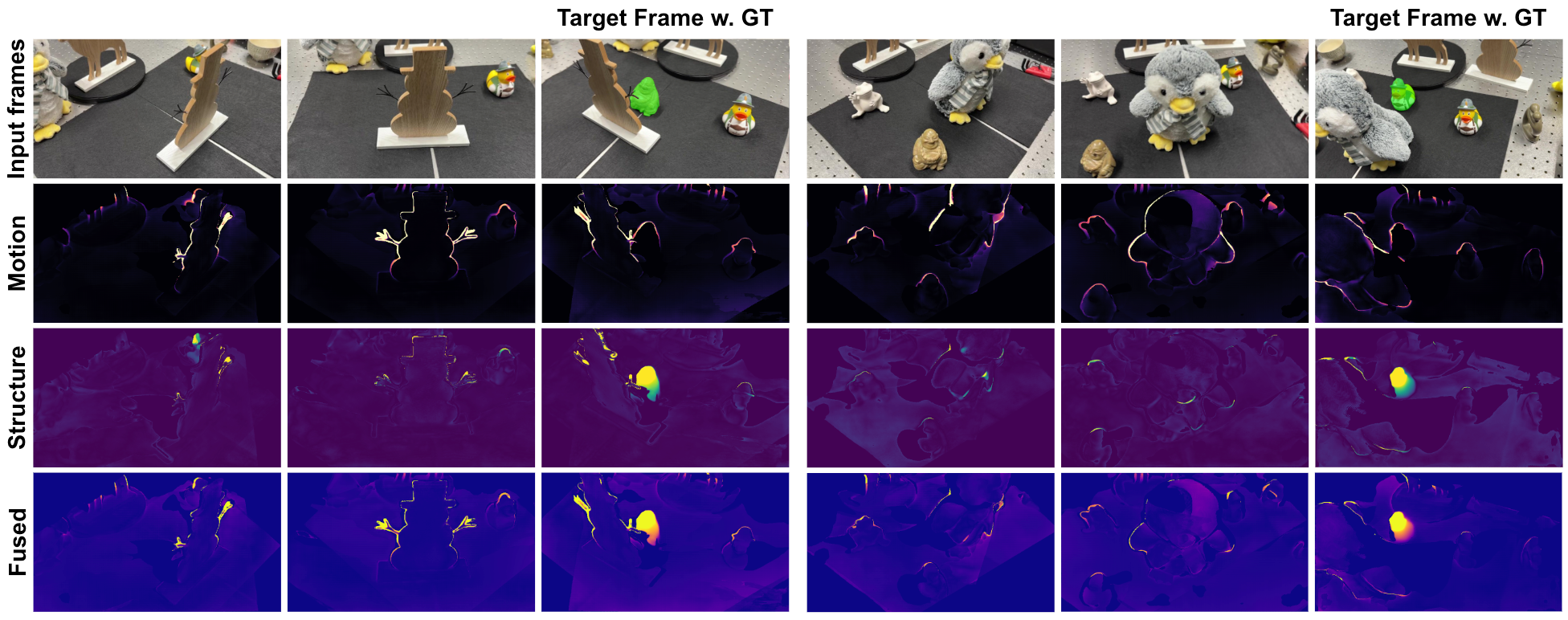}
\caption{
\textbf{Structure cue validation on OccluBench.}
We visualize two examples where objects appear abruptly in the target frame (Left: a statue; Right: a rubber duck). The structure map successfully highlights these inconsistencies. Conversely, the motion cue fails to capture these artifacts, as it classifies the appearing objects as occluded regions and consequently ignores the error region.
}
\label{fig:occlu_vis}
\end{figure}

\begin{figure}[tb]
\centering
    \includegraphics[width=\linewidth]{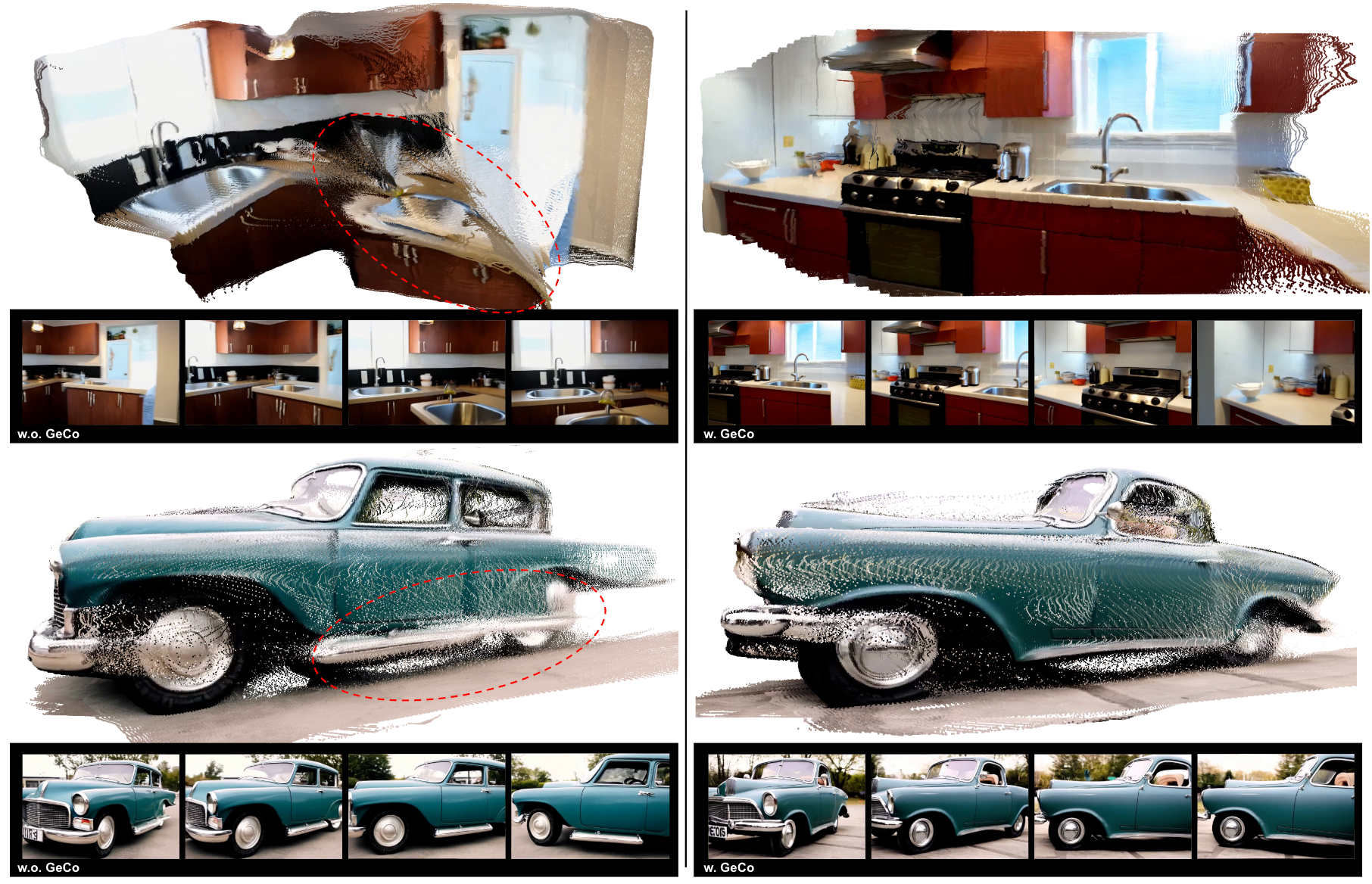}
\caption{
\textbf{Impact of \ourmethod{} guidance on reconstruction quality.} 
We observe that \ourmethod{} significantly reduces geometric artifacts compared to the baseline. 
Top: The baseline generates a sink that drifts over time (unnatural motion), leading to misalignment in the 3D reconstruction. 
Bottom: The baseline suffers from structural deformation, causing the car's side skirt to fracture into two disjoint segments.
}
\label{fig:supp_3d}
\end{figure}

\begin{figure}[!h]
\centering
    \includegraphics[width=\linewidth]{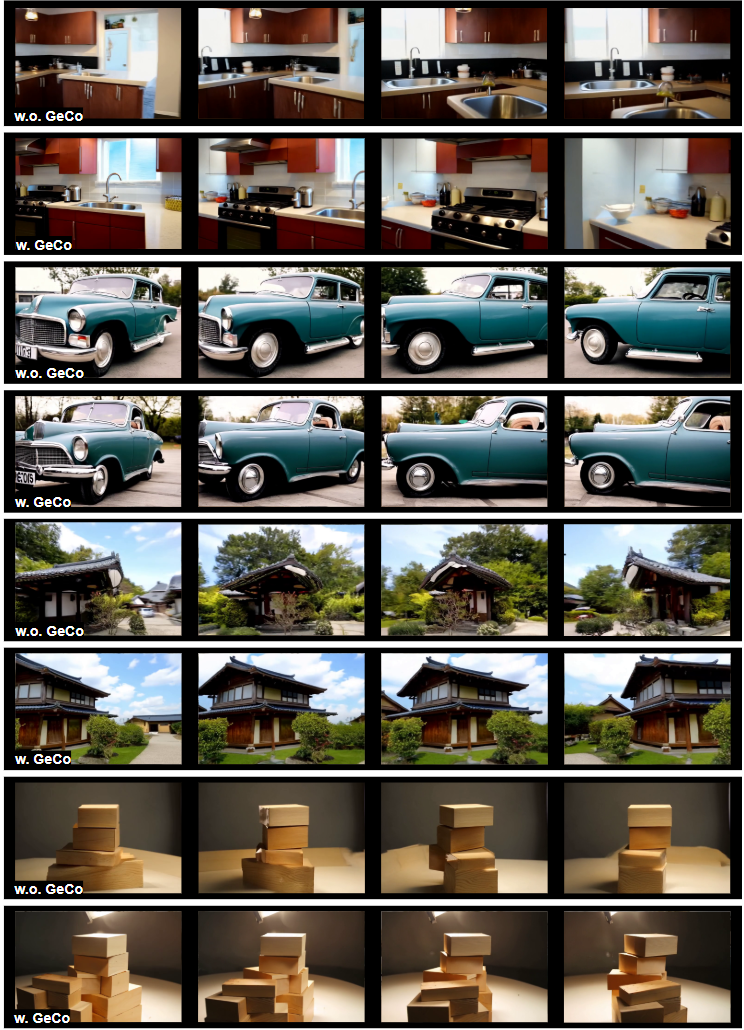}
\caption{\textbf{Extended visual comparison.} Videos generated with \ourmethod{} guidance consistently exhibit fewer deformation artifacts across diverse scenarios.}
\label{fig:supp_guidance}
\end{figure}

\noindent This supplementary material provides additional qualitative results and comprehensive implementation details to support the findings in the main paper.
We provide additional visual results (\cref{supp:more_qualitative}), details on structure consistency implementation, WarpBench development, and validation experiment setup (\cref{supp:implementation_detail}), hyperparameters for our training-free guidance experiments (\cref{sec:supp_hyperparams}), and details on \ourmethod{}-Eval benchmark (\cref{supp:eval_bench}) with the list of text prompts for evaluation (\cref{supp:full_list}). 


\section{More Qualitative Results}
\label{supp:more_qualitative}

\subsection{Interpretability Comparison}
\label{app:deformation_Qualitative}
Qualitative comparisons in \cref{fig:interpretability_comparison} highlight the utility of our metric as a diagnostic tool.
While baselines such as MEt3R~\cite{Asim2025MEt3RMM} produce blurred, non-specific score maps, \ourmethod{} generates precise maps that isolate specific geometric distortions, offering actionable feedback on model performance.
Under the rigid-scene assumption, non-rigid deformations appear as residual motion indistinguishable from independently moving objects. However, conventional motion-segmentation methods (\eg SegAnyMo~\cite{huang2025segment}) highlight the whole object area as deformation, showing ineffectiveness as a metric.

\subsection{Results on \ourmethod{} Validation Experiment}
We analyze the complementary nature of our motion and structure cues in Fig.~\ref{fig:warp_vis} and Fig.~\ref{fig:occlu_vis}. While the motion cue is essential for detecting surface deformations where depth remains consistent (Fig.~\ref{fig:warp_vis}), it often fails during sudden object appearances, misclassifying them as occlusions. In these latter cases, the structure cue proves critical, successfully identifying the hallucinated geometry that the motion cue ignores (Fig.~\ref{fig:occlu_vis}).

\subsection{Results on Guidance Experiment}

We provide an extended visual comparison of video generation quality in Fig.~\ref{fig:supp_guidance}. As shown in the figure, models employing \ourmethod{} guidance consistently maintain structural rigidity across complex motions, whereas the baselines exhibit noticeable deformation artifacts.

To further validate the geometric consistency of our generations, we perform 3D reconstruction on the generated clips. As illustrated in Fig.~\ref{fig:supp_3d}, videos generated with our guidance yield coherent point clouds, while baseline methods lead to reconstruction failures (\eg, object drift or fracturing) due to temporal inconsistencies.

\section{Implementation Details}
\label{supp:implementation_detail}

\subsection{Details on Structure Consistency}
\label{supp:structure_consistency}

In the main paper, we conceptualized structure consistency as the residual between the predicted depth and the reprojected depth map. In our implementation, we compute this using an inverse warping mechanism to ensure differentiability and sub-pixel accuracy via bilinear sampling.

Let $\mathbf{D}_c$ and $\mathbf{D}_i$ be the depth maps for the current (source) and reference (target) frames, respectively. For a pixel $\mathbf{p}$ in the current frame coordinate system with depth $z_c = \mathbf{D}_c(\mathbf{p})$, we back-project it to a 3D point $\mathbf{X}$ and project it onto the frame of $\mathbf{I}_i$:
\begin{equation}
    \mathbf{p}' = \mathbf{K}_i (\mathbf{R}_{c \to i} \mathbf{K}_c^{-1} \tilde{\mathbf{p}} z_c + \mathbf{t}_{c \to i}),
\end{equation}
where $\tilde{\mathbf{p}}$ is the homogeneous coordinate of $\mathbf{p}$. Let $z_{\text{proj}}$ be the z-coordinate of the point in the reference frame (the predicted distance to the camera).

We sample the reference depth map at the projected location $\mathbf{p}'$ using bilinear interpolation to obtain $\hat{d}_i = \mathbf{D}_i(\mathbf{p}')$. The structure consistency error at pixel $\mathbf{p}$ is computed as the normalized relative depth difference:
\begin{equation}
    \mathcal{L}_{\text{geo}}(\mathbf{p}) = \frac{\left| \hat{d}_i - z_{\text{proj}} \right|}{z_c}.
\end{equation}

\paragraph{Occlusion Handling.}
Instead of explicit z-buffering, we employ a bi-directional geometric consistency check to generate a validity mask $\mathbf{M}_{\text{vis}}$. A pixel $\mathbf{p}$ is considered valid (co-visible) only if it satisfies the forward visibility check:
\begin{equation}
    \frac{z_{\text{proj}} - \hat{d}_i}{z_{\text{proj}}} < \tau,
\end{equation}
where $\tau$ is a relative threshold (set to 0.02 in experiments). This filters out pixels where the source point is occluded by a closer surface in the target view. We further refine this mask by performing the inverse check (projecting from $i$ to $c$) and retaining only the intersection of valid regions.

\subsection{Details on Thin-plate-spline for WarpBench}
As described in the main paper, we develop WarpBench based on two open-source datasets: CO3D~\cite{reizenstein21co3d} (\textit{CO3D-Warp}) with object-centric clips and ScanNet++~\cite{dai2017scannet,yeshwanthliu2023scannetpp} (\textit{ScanNet-Warp}) with indoor scenes clips. In total, \textsc{WarpBench} contains 100 object-centric clips (2{,}000 frames) and 100 scene-level clips (2{,}000 frames), spanning 50 object categories and 6 indoor environments.

To simulate deformation artifacts, we inject temporally smooth non-rigid warps parameterized by thin-plate splines (TPS).
For each clip, we sample $K$ control points $c_i$ within the foreground mask via farthest-point sampling (FPS) to ensure coverage. Their 2D displacements $\mathbf{u}_{i,t} \in \mathbb{R}^2$ evolve under a simple temporal model. At each frame $t$, we fit a TPS warp that maps fixed control points $c_i$ to targets $y_{i,t}$:
\begin{equation}
    y_{i,t} = c_i + \mathbf{u}_{i,t}.
\end{equation}

Specifically, we model the warp as an affine-plus-RBF mapping $f_t: \mathbb{R}^2 \!\to\! \mathbb{R}^2$ with a TPS basis $\phi(r) = r^2 \log r$. The mapping is defined as:
\begin{equation}
    f_t(x) \;=\; A_t x + a_t + \sum_{i=1}^K w_{i,t}\,\phi(\|x-c_i\|),
\end{equation}
where $A_t \in \mathbb{R}^{2\times 2}$, $a_t \in \mathbb{R}^2$, and $w_{i,t} \in \mathbb{R}^2$ are the TPS parameters. The dense forward displacement $U_t(p)$ at pixel $p$ is derived by:
\begin{equation}
    U_t(p) = f_t(p) - p.
\end{equation}

We localize the deformation with a feathered weight map $w(p) \in [0,1]$ to calculate $\tilde U_t(p)$, and subsequently enforce temporal smoothness via an exponential moving average $\bar U_t(p)$:
\begin{align}
    \tilde U_t(p) &= w(p)\,U_t(p), \\
    \bar U_t(p) &= \beta\,\bar U_{t-1}(p) + (1-\beta)\,\tilde U_t(p).
\end{align}

The final warped frame is obtained by differentiable backward sampling:
\begin{equation}
    I^{\mathrm{def}}_t(p) = I_t\big(p + \bar U_t(p)\big).
\end{equation}

For every warped frame, we release the dense displacement field $\bar U_t(p) \in \mathbb{R}^2$ as ground truth, along with its magnitude $M_t(p)$ when a scalar target is needed:
\begin{equation}
    M_t(p) = \|\bar U_t(p)\|_2.
\end{equation}

\subsection{Spearman’s Rank Correlation Coefficient}
In the deformation localization experiment of WarpBench, we used Spearman’s rank correlation coefficient (SRCC) to measure rank correlation with the ground truth deformation magnitude, here we provide a detailed explanation of this metric: SRCC is a non-parametric measure of rank correlation that evaluates the strength and direction of a monotonic relationship between two ranked variables, defined as $\rho = 1 - \tfrac{6 \sum d_i^2}{n(n^2 - 1)}$, where $d_i$ is the difference between the ranks of paired observations and $n$ is the number of pairs.

\subsection{Validation Experiment Setup}
\label{sec:supp_validation}

\myparagraph{Synthetic validation (TartanAir~v2).}
We evaluate on the TartanAir~v2~\cite{Wang2020TartanAirAD} dataset using the \texttt{lcam\_front} camera.
For each scene we process both the \emph{Easy} and \emph{Hard} difficulty splits.
We randomly sample up to $k{=}50$ consecutive frame pairs per trajectory (seed $42$), yielding approximately 10k frame pairs in total.
Predicted depths are aligned to the ground-truth scale via median scaling before computing errors.
Valid-depth pixels are restricted to the range $[2,\,40]$\,m; the covisibility confidence threshold is set to $\tau_\text{cov}{=}0.5$.

\myparagraph{Real-world noise floor (DL3DV).}
We additionally run \ourmethod{} on 30 scenes from DL3DV~\cite{Ling2023DL3DV10KAL}, a dataset of real-world videos with predominantly rigid camera motion, using the same backbone models and evaluation parameters as above.
Because no ground-truth geometry is available, only predicted scores are reported; the resulting fused score serves as an empirical real-world noise floor.

\myparagraph{Component-level metrics.}
For each frame pair we report three backbone error metrics against ground truth:
(i)~optical flow end-point error (EPE, in pixels),
(ii)~depth absolute relative error (AbsRel, after median-scale alignment), and
(iii)~camera pose angular error, computed as the geodesic rotation error plus the translation direction error (in degrees) from the relative pose between the two frames.

\myparagraph{Sensitivity to input noise.}
To stress-test sensitivity, we inject controlled Gaussian noise independently into the \emph{ground-truth} flow and depth of the TartanAir~v2 pairs (200 randomly sampled pairs across both splits).
Flow is perturbed additively in pixel space:
\begin{equation}
    \mathbf{f}_\text{noisy}(i,j) = \mathbf{f}_\text{GT}(i,j) + \boldsymbol{\epsilon}, \quad \boldsymbol{\epsilon} \sim \mathcal{N}(\mathbf{0},\, \sigma_\text{flow}^2 \mathbf{I}),
\end{equation}
where $\sigma_\text{flow} \in \{0, 0.25, 0.5, 1, 2, 3, 5, 8, 12, 20\}$~pixels.
Depth is perturbed multiplicatively in log-space so that errors scale proportionally with distance:
\begin{equation}
    D_\text{noisy} = D_\text{GT} \cdot \exp(\epsilon), \quad \epsilon \sim \mathcal{N}(0,\, \sigma_\text{depth}^2),
\end{equation}
where $\sigma_\text{depth} \in \{0, 0.01, 0.02, 0.05, 0.1, 0.15, 0.2, 0.3, 0.5, 1.0\}$.
In each sweep only one input is corrupted while all others (depth/camera or flow, respectively) remain at ground truth, isolating each component's contribution.
The backbone prediction operating points---$\text{EPE} = 0.757/1.155$~px for flow and $\text{AbsRel} = 0.056/0.061$ for depth on Easy/Hard---are marked on each plot for reference, alongside the typical generated-video fused score range.

\section{Hyperparameters for Training-free Guidance Experiment}
\label{sec:supp_hyperparams}

We conduct our experiments using the CogVideoX-5B text-to-video model as the base generator. All inference is performed using \texttt{bfloat16} precision on a single NVIDIA H200 GPU. The generated videos consist of $N=49$ frames with a resolution consistent with the standard model output.

\subsection{\ourmethod{} Guidance Configuration}
To compute the geometric consistency loss $\mathcal{L}_{\text{geo}}$, we utilize two frozen pretrained backbones: VGGT-1B~\cite{wang2025vggt} for depth and camera estimation, and UFM-Base~\cite{zhang2025ufm} for optical flow. 

To ensure computational efficiency, we do not decode the full sequence for every gradient update. Instead, we apply Latent Slicing~\cite{Jang2025FrameGT} on a fixed set of anchor frames, specifically indices $\mathcal{I} = \{12, 24, 36, 48\}$ (1-based indexing). The residual motion metric is computed in \textit{adjacent} pair mode ($k \to k+1$) across these frames. To further accelerate the pipeline, input frames passed to the flow network are downscaled by a factor of $0.5$.

\subsection{Sampling and Optimization Schedule}
We utilize a standard DDIM sampling trajectory with $T=50$ inference steps and a Classifier-Free Guidance (CFG) scale of $6.0$. 

Following the optimization strategy proposed in~\cite{Jang2025FrameGT}, we apply a recursive denoising schedule where the number of gradient updates $R_t$ varies per timestep $t$. We utilize a constant step size (learning rate) of $\eta = 3.0$ for all updates. The schedule is divided into three distinct phases:
\begin{enumerate}
    \item Warm-up ($t \in [0, 2]$): We perform no gradient updates ($R_t=0$) in the initial steps to establish the global layout.
    \item Strong Guidance ($t \in [3, 19]$): We apply $R_t=3$ updates per step to enforce strong geometric constraints during the formation of structural content.
    \item Refinement ($t \in [20, 49]$): We reduce the frequency to $R_t=2$ updates per step to maintain consistency without disrupting fine texture generation.
\end{enumerate}

To mitigate the accumulation of errors and prevent the latent from drifting off the data manifold during aggressive updates, we strictly employ Time-Travel~\cite{He2023ManifoldPG} within the specific interval $t \in [15, 20]$.

\section{Details on \ourmethod{}-Eval Benchmark}
\label{supp:eval_bench}

\subsection{Model Configurations}
Tab.~\ref{tab:model-fps} summarizes the native frame rate of each video generation model evaluated in our benchmark.

\subsection{Prompts Design}
As described in the main paper, we design the benchmark to include four scenarios: 
\emph{(i)} object-centric scenes, \emph{(ii)} indoor scenes, \emph{(iii)} outdoor scenes, and \emph{(iv)} appearance-complex scenes (\eg, characterized by high-frequency patterns, reflections, and thin structures). We show the aggregated summary of the prompts in Table~\ref{tab:task_benchmark_final}, and the full list of the prompts in \cref{supp:full_list}.

\begin{table}[tb]
\small
\caption{\textbf{Frame rate of evaluated models.}}
\label{tab:model-fps}
\resizebox{\linewidth}{!}{%
\begin{tabular}{@{}l *{8}{c} @{}}
\toprule
Model & Sora 2 & Veo 3.1 & WAN 2.2 & HunyuanVideo & CogVideoX1.5 & CogVideoX-5B & CogVideoX-2B & LTX-Video \\
\midrule
Type & \cellcolor{gray!12}Commercial & \cellcolor{gray!12}Commercial & Open-Source & Open-Source & Open-Source & Open-Source & Open-Source & Open-Source \\
FPS  & 30 & 24 & 16 & 24 & 16 & 16 & 16 & 30 \\
\bottomrule
\end{tabular}%
}
\end{table}

\begin{table}[tb]
\centering
\footnotesize
\caption{\textbf{Composition of the \ourmethod{}-Eval benchmark}. The benchmark is structured around four evaluation scenarios designed to probe geometric consistency, each tested with both slow and fast camera dynamics.}
\label{tab:task_benchmark_final}
\begin{tabularx}{\linewidth}{@{} p{0.3\linewidth} l X @{}}
\toprule
Evaluation Scenario & Dynamics & Example Scenarios \\
\midrule
\multirow{2}{=}{\parbox{\linewidth}{
\vspace{1em}
\textbf{Object-centric scenes}: Maintain rigid, detailed geometry of a single isolated object.}}
 & Slow & 360° orbit of marble bust; push-in on antique globe; upward tilt on bronze horse statue. \\
 \cmidrule(l){2-3}
 & Fast & Rapid orbit of espresso machine; fast dolly in/out on jeweled pocket watch; inward spiral around stone gargoyle. \\
\midrule
\multirow{2}{=}{\parbox{\linewidth}{
\vspace{0.4em}
\textbf{Indoor scenes}: Preserve coherent room layout and global structure during traversal.}}
 & Slow & Grand library aisle dolly; gallery arc around abstract sculpture; server-aisle glide. \\
 \cmidrule(l){2-3}
 & Fast & Office hallway sprint; boiler-room right-angle turns; spiral staircase ascent. \\
\midrule
\multirow{2}{=}{\parbox{\linewidth}{
\vspace{1em}
\textbf{Outdoor scenes}: Ensure consistency across expansive environments with layered depth.}}
 & Slow & Cloister courtyard traverse; under-bridge arch glide; shipping-container row track. \\
 \cmidrule(l){2-3}
 & Fast & Parking-structure sharp turns; underpass colonnade weave; container maze with two right-angle turns. \\
\midrule
\multirow{2}{=}{\parbox{\linewidth}{
\vspace{0.5em}
\textbf{Appearance-complex scenes}: Test fine patterns, reflections, refractions, and dense edges prone to artifacts.}}
 & Slow & Chrome motorcycle orbit (stable reflections); glass-brick refraction dolly; mosaic floor lateral slide. \\
 \cmidrule(l){2-3}
 & Fast & Mirror-room orbit; fast lateral across venetian blinds; weave through beaded glass curtains. \\
\bottomrule
\end{tabularx}
\end{table}

\subsection{Motion Metrics}
Models that generate very little motion can appear highly temporally consistent, even when they ignore motion-rich prompts. Prior work therefore uses optical flow to detect near-static generations: \emph{EvalCrafter}~\cite{Liu2023EvalCrafterBA} defines a Flow-Score as the clip-averaged flow magnitude and a Motion AC-Score by thresholding this value, while \emph{VBench/VBench++}~\cite{Huang2023VBenchCB,Huang2024VBenchCA} report a Dynamic Degree based on whether the largest $5\%$ of per-frame flow magnitudes exceed a threshold. These metrics, however, are not normalized for resolution/FPS and are largely threshold-based, making cross-model comparison difficult when generation settings differ.

Hence, we use a continuous motion statistic with resolution/FPS normalized. Given frames $I_t$ ($t \in \{1,\dots,T$\}), we estimate the optical flow $F_t(x,y)$ and its magnitude $M_t(x,y)$ between $I_t$ and $I_{t+1}$:
\begin{align}
    F_t(x,y) &= (u_t, v_t), \\
    M_t(x,y) &= \sqrt{u_t^2+v_t^2}.
\end{align}
Let $W,H$ be the image width and height. We normalize motion relative to the image diagonal $D=\sqrt{W^2+H^2}$. After masking invalid pixels, we compute the per-pair normalized displacement $m_t$:
\begin{equation}
    m_t = \operatorname{mean}_{x,y}\left[ \frac{M_t(x,y)}{D} \right].
\end{equation}
Finally, for a clip of duration $S$ seconds, we define the Total Motion ($\mathrm{TM}$) and Motion Intensity ($\mathrm{MI}$) as:
\begin{align}
    \mathrm{TM} &= \sum_{t=1}^{T-1} m_t, \\
    \mathrm{MI} &= \frac{\mathrm{TM}}{S}.
\end{align}
Here, $\mathrm{TM}$ is a dimensionless measure (fraction of the image diagonal) representing the total path length, which is approximately invariant to FPS for a fixed $S$. $\mathrm{MI}$ represents the corresponding average motion per second. In \ourmethod{}-Eval, we report per-model, per-scenario means and standard deviations of $\mathrm{TM}$ and $\mathrm{MI}$, allowing us to interpret geometric consistency jointly with the magnitude of motion produced by the model.

\newpage
\clearpage
\onecolumn

\section{List of Text Prompt}
\label{supp:full_list}
\providecommand{\bftext}[1]{\textbf{#1}}

\subsection{Object-Centric Prompts}
\subsubsection{Slow camera movements}

\noindent\bftext{a1.} In a serene gallery, a slow, deliberate 360-degree orbit captures a masterfully carved marble bust, poised elegantly on a polished pedestal. The bust, depicting a serene figure with intricate details, remains perfectly centered as the camera glides smoothly around it. The lighting casts soft shadows, accentuating the delicate features and the smooth texture of the marble. The exposure and white balance are meticulously set, ensuring the bust's timeless beauty is captured in pristine clarity. The focus remains sharp, highlighting every curve and chisel mark, while the background fades into a gentle blur, emphasizing the sculpture's artistry. The only motion is the camera's graceful orbit, creating a mesmerizing, uninterrupted view of this exquisite masterpiece.

\noindent\bftext{a2.} The camera gracefully advances toward an exquisite antique globe perched on an intricately carved wooden stand, its surface illuminated by soft, ambient light. As the frame narrows, the globe's aged paper grain becomes evident, showcasing faded cartographic details and muted colors that whisper tales of exploration. The brass meridian, polished yet timeworn, gleams subtly, its engraved lines and numbers hinting at a bygone era of navigation. The room is enveloped in a serene stillness, with no hint of wind or vibration, allowing the globe's historical elegance to captivate the viewer. The camera's gentle motion is the only dynamic element, drawing the eye closer to the globe's timeless allure.

\noindent\bftext{a3.} The video begins with a close-up of the bronze horse statue's hooves, capturing the intricate details of the sculpted muscles and the polished sheen of the metal. As the camera tilts upward at a steady, deliberate pace, the viewer is drawn into the artistry of the statue, revealing the powerful legs and the graceful curve of the horse's body. The room remains still and silent, with soft, even lighting casting gentle shadows that accentuate the statue's form. The focus remains sharp, highlighting the texture and craftsmanship of the bronze. The camera continues its ascent, passing the horse's strong neck, until it reaches the noble head, capturing the lifelike expression and finely detailed features, completing the serene and contemplative journey.

\noindent\bftext{a4.} A meticulously crafted lateral dolly shot begins at the platen of a vintage typewriter, its textured roller and paper guide evoking a sense of nostalgia. The camera glides smoothly, revealing the intricate array of keys, each with its own unique patina, hinting at countless stories typed over the years. The levers and typebars, though stationary, suggest potential energy, poised for action. As the camera continues its journey, the focus shifts to the side panel, showcasing the elegant curves and mechanical artistry of the typewriter's design. The entire scene is bathed in soft, ambient light, highlighting the timeless beauty of this classic writing instrument.

\noindent\bftext{a5.} The camera begins its graceful descent from a bird's-eye view, revealing a stunning ceramic vase with intricate glazed relief patterns. As it slowly lowers, the vase's exquisite details become more pronounced, showcasing delicate floral motifs and swirling designs in rich, earthy tones. The vase sits on a polished wooden surface, its glossy finish reflecting the soft ambient light. The surroundings remain perfectly still, enhancing the vase's elegance and craftsmanship. The camera continues its descent, eventually settling into a tight side profile, capturing the vase's curves and textures with precision. The lighting remains constant, casting gentle shadows that accentuate the vase's artistry, creating a serene and contemplative visual experience.

\noindent\bftext{a6.} The camera gracefully arcs around a luxurious gilded wall mirror, its ornate frame glistening with intricate golden details. The mirror reflects a serene, softly lit room, capturing only the immediate backdrop—a plush velvet armchair and a delicate vase of fresh lilies on a polished wooden table. The camera maintains a consistent distance, ensuring the reflections remain steady and undisturbed. The room's ambiance is tranquil, with warm, ambient lighting casting gentle shadows. The mirror's surface is pristine, reflecting the elegance of the setting without any flicker or change, as the camera's smooth motion reveals the timeless beauty of the scene.

\noindent\bftext{a7.} The camera begins its slow, deliberate journey at the base of a towering wooden abstract sculpture, its intricate grain and texture immediately captivating. As the dolly glides diagonally upward, the sculpture's complex curves and angles are revealed, each facet catching the ambient light in a unique way, creating a play of shadows and highlights. The fixed focus ensures every detail remains sharp, while the consistent white balance maintains the sculpture's natural hues. The background remains unobtrusive, allowing the sculpture's artistry to dominate the frame. The camera's smooth ascent culminates at the sculpture's opposite top corner, offering a final, breathtaking perspective of this wooden masterpiece.

\noindent\bftext{a8.} The camera glides smoothly in a precise orbit around an ancient stone relief wall panel, capturing the intricate artistry of the chiseled figures. As it moves laterally, the depth of each groove and the play of light and shadow in the recessed areas become more pronounced, revealing the skill of the artisans. The relief, depicting a serene scene of historical significance, remains perfectly still, allowing the viewer to appreciate the fine details and textures. The environment, a dimly lit gallery with soft ambient lighting, enhances the timeless quality of the stone, while the camera's path is the sole source of motion, creating a dynamic yet tranquil viewing experience.

\noindent\bftext{a9.} A luxurious mechanical pocket watch rests open on deep crimson velvet, its intricate details captured in stunning macro. The camera begins its slow, deliberate pan at the ornate crown, highlighting the delicate engravings and polished metal. As the lens glides smoothly across the watch's surface, the frozen hands and gears are revealed, each component meticulously crafted and gleaming under the soft, focused illumination. The journey continues to the escapement, where the precision of the jewel settings is showcased, their vibrant colors contrasting beautifully with the metallic intricacies. The entire scene exudes timeless elegance, with the velvet's rich texture enhancing the watch's exquisite craftsmanship.

\noindent\bftext{a10.} The camera gracefully orbits a meticulously crafted bonsai tree, positioned on a low, elegant stand, capturing the intricate details of its textured bark and the delicate wirework shaping its branches. The lens remains at leaf height, offering an intimate view of the bonsai's artistry, with each leaf and branch in sharp focus. The serene atmosphere is undisturbed by any breeze, ensuring the leaves remain perfectly still, enhancing the tranquility of the scene. The exposure is expertly balanced, highlighting the rich hues of the bonsai's foliage and the subtle variations in its bark. The camera's smooth, continuous motion is the sole dynamic element, creating a mesmerizing visual experience.

\subsubsection{Fast camera movements}

\noindent\bftext{a11.} The camera swiftly orbits a gleaming, polished metal espresso machine, capturing its sleek, reflective surface in stunning detail. As it circles, the machine's curves and edges reveal dazzling specular highlights, creating a dynamic play of light and shadow. The environment remains perfectly still, with the espresso machine standing as the centerpiece against a minimalist backdrop. The lighting is expertly set, casting soft, even illumination that accentuates the machine's contours. The camera's rapid movement provides a seamless 360-degree view, showcasing the espresso machine's elegant design and craftsmanship from every angle, while the surroundings remain unchanged.

\noindent\bftext{a12.} The video begins with a swift dolly-in towards an exquisite, vintage pocket watch, its surface adorned with intricate jewels that catch the light. As the camera zooms in, the focus sharpens on the delicate, engraved monogram, revealing the initials "J.L." in elegant script. The watch's face, a masterpiece of craftsmanship, is frozen in time, its hands paused at precisely 10:10. The exposure remains constant, highlighting the watch's gleaming metallic finish and the sparkle of its jewels. Suddenly, the camera rapidly pulls back, revealing the full grandeur of the pocket watch, resting on a plush velvet surface, its timeless elegance captured in perfect clarity.

\noindent\bftext{a13.} The video begins with a close-up of the base of a grand, fluted stone column, its intricate grooves and textures highlighted by steady, soft lighting. The camera swiftly tilts upward, maintaining perfect alignment with the column's vertical lines, revealing the elegant, timeless craftsmanship of the stonework. As the camera ascends, the column's fluting creates a mesmerizing pattern, leading the viewer's eye toward the capital. Upon reaching the top, the capital is revealed in all its ornate glory, adorned with classical acanthus leaves and scrolls, set against the backdrop of a stately, immobile room. The entire scene remains serene and unchanging, with the brisk camera movement providing the only sense of motion.

\noindent\bftext{a14.} A sleek, dynamic arc swiftly encircles a vintage film camera perched on a sturdy tripod, capturing its essence from a front three-quarter angle to a rear three-quarter perspective. The camera, a classic piece with intricate dials and a polished lens, remains perfectly still, exuding timeless elegance. The tripod, robust and unwavering, supports the camera with precision. As the viewpoint sweeps rapidly around, maintaining a constant height, the background blurs into a soft focus, emphasizing the camera's intricate details and craftsmanship. The lighting casts subtle highlights on the camera's metallic surface, enhancing its vintage allure.

\noindent\bftext{a15.} A sleek, dynamic shot captures a rapid lateral pass across a classic wooden chessboard, where two rooks stand as sentinels at opposite ends. The camera glides swiftly, skimming just above the meticulously arranged pieces, each casting a sharp shadow on the polished squares. The board's plane remains nearly parallel to the sensor, creating a mesmerizing blur of alternating dark and light squares. The pieces, carved with precision, stand immobile and unwavering, their intricate details highlighted by steady, soft lighting. The focus remains sharp, emphasizing the rooks' commanding presence, while the camera's swift motion creates a sense of urgency and fluidity.

\noindent\bftext{a16.} The camera swiftly ascends from a polished wooden tabletop, capturing the elegant rise towards a crystal decanter, its intricate facets glistening under soft, ambient lighting. The room, adorned with rich mahogany furniture and subtle, warm tones, remains perfectly still, exuding an air of timeless sophistication. As the camera reaches its zenith, the decanter's pristine surface reflects the light in a steady, mesmerizing dance of highlights, maintaining a serene, unwavering brilliance. The final top-down view reveals the decanter's symmetrical beauty, its contents a deep amber, contrasting with the room's muted elegance, creating a harmonious visual symphony.

\noindent\bftext{a17.} The camera swiftly pans across a meticulously arranged row of antique bottles, each uniquely shaped and colored, their glass surfaces reflecting a warm, ambient light. As the camera moves, the labels blur into a kaleidoscope of vintage typography and faded hues. Suddenly, the motion halts with precision on the central bottle, its label featuring ornate script and intricate detailing, capturing the essence of a bygone era. The camera lingers momentarily, allowing the viewer to absorb the craftsmanship and history encapsulated in the label. Then, with a seamless transition, the camera resumes its rapid journey, leaving the bottles in a serene, undisturbed stillness.

\noindent\bftext{a18.} The camera begins its journey with a swift upward motion, revealing a vintage tabletop gramophone in exquisite detail. As it ascends, the polished wooden base and intricate brass horn come into view, capturing the essence of a bygone era. The camera then gracefully arcs over the gramophone, offering a bird's-eye perspective of the stationary needle poised above the silent vinyl record. The lighting casts gentle shadows, enhancing the gramophone's timeless elegance. As the camera descends on the opposite side, it focuses on the record's grooves, capturing the texture and craftsmanship. The scene concludes with the camera at platter height, emphasizing the gramophone's silent anticipation, as if waiting for the music to begin.

\noindent\bftext{a19.} The camera swiftly zig-zags towards a charming, intricately detailed miniature dollhouse, capturing its quaint architecture and tiny furnishings in high definition. The first quick step reveals the dollhouse's vibrant exterior, with its pastel colors and delicate window frames, while the second step shifts the perspective, offering a closer view of the meticulously arranged furniture inside, including a tiny wooden table and chairs. Throughout the sequence, the horizon remains perfectly level, emphasizing the dynamic movement of the camera against the stillness of the scene. The absence of any breeze or motion within the dollhouse enhances the contrast between the camera's rapid approach and the serene, static environment.

\noindent\bftext{a20.} The camera begins its journey with a sweeping motion around a majestic stone gargoyle perched atop an ancient cathedral, its intricate details illuminated by the soft glow of twilight. As the camera spirals inward, the gargoyle's menacing features, including its sharp fangs and piercing eyes, become more pronounced against the backdrop of the dusky sky. The spiral tightens smoothly, revealing the texture of the weathered stone, capturing every crack and crevice with precision. The background remains a constant, blurred tapestry of gothic architecture, while the exposure and focus remain locked, ensuring the gargoyle's fierce expression is the focal point. The spiral concludes with a tight profile shot, emphasizing the gargoyle's formidable presence and the artistry of its creation.

\subsection{Indoor Prompts}
\subsubsection{Slow camera movements}

\noindent\bftext{b1.} The camera glides smoothly down the central aisle of a grand library, flanked by towering bookcases packed tightly with an array of colorful books, their spines creating a mosaic of knowledge. The shelves, lined with neatly arranged volumes, are interspersed with small, elegant signs indicating various genres and sections. The lighting is warm and steady, casting a gentle glow that highlights the rich wood of the bookcases and the intricate patterns of the carpeted floor. As the camera advances at a constant pace, the scene remains perfectly still, evoking a sense of timelessness and tranquility, inviting viewers to immerse themselves in the serene ambiance of this literary sanctuary.

\noindent\bftext{b2.} In a serene, minimalist gallery, the camera gracefully arcs around a solitary abstract sculpture, its smooth curves and intricate textures highlighted by the steady, soft lighting. The sculpture, a fusion of metal and stone, stands as the focal point, its form evoking a sense of mystery and contemplation. Glass display cases, housing delicate artifacts, and subtle wall labels linger in the periphery, their presence understated yet integral to the gallery's ambiance. The constant lighting casts gentle shadows, enhancing the sculpture's allure. The only movement is the camera's fluid motion, creating a tranquil, immersive experience that invites reflection and appreciation.

\noindent\bftext{b3.} The camera glides smoothly down an opulent hotel corridor, perfectly centered, revealing a symphony of intricate details. Richly patterned carpets stretch endlessly beneath, their designs echoing elegance and sophistication. Ornate sconces cast a warm, steady glow, illuminating the corridor with a timeless ambiance. Each door, identical yet unique, stands as a sentinel, adorned with polished brass numbers and gleaming handles. Discreet signage, tastefully placed, guides unseen guests with understated grace. The walls, adorned with subtle textures, frame the scene, while the camera's deliberate, unwavering motion creates a mesmerizing journey through this serene, luxurious passageway.

\noindent\bftext{b4.} The camera glides smoothly across a pristine stainless-steel commercial kitchen, capturing the gleaming surfaces of industrial-grade appliances, neatly arranged shelves, and an array of utensils and pans. The countertops, lined with culinary tools, reflect the ambient light, creating a serene, almost clinical atmosphere. As the camera moves parallel to the counters, the symmetry of the kitchen's layout becomes apparent, with each element meticulously organized. The fixed exposure and focus highlight the kitchen's immaculate condition, emphasizing the polished metal and the quiet stillness of the space, devoid of any steam, water, or fan motion, offering a tranquil, undisturbed view.

\noindent\bftext{b5.} In a grand museum hall, the camera glides gracefully, capturing the serene ambiance. Display plinths, adorned with intricate artifacts, stand in orderly rows, each accompanied by detailed information placards. The lighting casts a soft, consistent glow, highlighting the exhibits' historical significance. As the camera moves, it reveals the majestic dinosaur skeleton, its massive feet firmly planted, leading up to its towering skull. The skeletal structure, a testament to ancient times, remains motionless, exuding a sense of timelessness. The camera's steady drift emphasizes the hall's quiet reverence, inviting viewers to immerse themselves in the museum's rich tapestry of history.

\noindent\bftext{b6.} The camera glides smoothly along a narrow server aisle, maintaining a precise 1.5-meter height, capturing the dense arrangement of towering server racks. Each cabinet is meticulously organized, with neatly labeled cables running in disciplined lines, creating a sense of order and efficiency. The steady glow of indicator lights on the servers casts a soft, ambient illumination, highlighting the sleek, metallic surfaces of the equipment. The environment is silent and unyielding, exuding a sense of technological precision and control. As the camera moves parallel to the cabinets, the rigid symmetry of the scene is emphasized, offering a glimpse into the heart of a meticulously maintained data center.

\noindent\bftext{b7.} The camera smoothly glides down a bustling supermarket aisle, capturing a vibrant array of neatly stacked boxed and canned goods, each adorned with colorful labels and enticing graphics. Price tags dangle from the shelves, offering deals and discounts, while shelf talkers highlight special promotions, all frozen in time. The camera maintains a shallow oblique angle, providing a dynamic perspective that emphasizes the abundance and variety of products. The aisle is brightly lit, casting a warm glow over the scene, as the camera advances steadily, creating a sense of anticipation and discovery amidst the stillness of the packaging and signage.

\noindent\bftext{b8.} In a spacious, modern conference room, the camera smoothly glides around a large, polished wooden table, meticulously set for a meeting. The table is surrounded by sleek, ergonomic chairs, each with a notepad and pen neatly placed in front. Crystal-clear water glasses catch the light, casting subtle reflections on the table's surface. Cable grommets are strategically positioned, hinting at the room's technological readiness. The ambient lighting is soft and consistent, creating a professional yet inviting atmosphere. As the camera circles, the scene remains perfectly still, capturing the anticipation of a gathering, with only the camera's gentle motion breaking the serene stillness.

\noindent\bftext{b9.} The camera glides slowly through a dimly lit antique shop, revealing a labyrinth of stacked vintage furniture, ornate lamps, and gilded frames. Each shelf overflows with eclectic trinkets, from delicate porcelain figurines to tarnished brass compasses. The air is thick with the scent of aged wood and history, as the camera weaves between narrow aisles, brushing past velvet-upholstered chairs and intricately carved tables. Dust motes dance in the soft, golden light filtering through stained glass windows, casting colorful patterns on the worn wooden floor. The shop remains eerily still, a silent guardian of forgotten stories, as the camera's gentle motion breathes life into the timeless treasures.

\noindent\bftext{b10.} The camera smoothly glides through a bustling workshop, revealing a series of workbenches laden with an array of hand tools, neatly organized bins, and pegboards adorned with hanging implements. Parts trays, filled with various components, line the benches, each meticulously arranged. The lighting casts a warm, consistent glow, highlighting the rich textures of wood and metal. The focus remains sharp, capturing every detail of the stationary objects, from the gleaming wrenches to the colorful bins. As the camera tracks parallel to the benches, the scene exudes a sense of order and craftsmanship, with the fixed exposure maintaining a clear, vivid view of this industrious space.

\subsubsection{Fast camera movements}

\noindent\bftext{b11.} The camera races down an endless office hallway, capturing a thrilling high-speed journey. The walls are lined with identical wooden doors, each with polished brass handles, creating a rhythmic pattern. Bright red exit signs hang above every few doors, their glow casting a steady light on the neutral-toned walls. Overhead, fluorescent ceiling fixtures illuminate the path with a consistent, cool white light, casting no shadows. The floor is carpeted in a muted gray, providing a sense of continuity. As the camera zooms forward, the static objects blur slightly at the edges, enhancing the sensation of speed, while maintaining perfect alignment and balance.

\noindent\bftext{b12.} The camera glides swiftly through a vast warehouse aisle, flanked by towering pallet racks stacked with neatly arranged cartons, each adorned with vibrant labels. The scene is bathed in soft, ambient lighting, casting gentle shadows on the polished concrete floor. As the camera hugs the centerline, it executes smooth, fluid yaw changes, offering dynamic perspectives of the meticulously organized inventory. The pallets and signage remain perfectly still, creating a striking contrast to the seamless motion of the camera. The journey through the aisle feels like a dance, with the camera's graceful movements highlighting the warehouse's orderly precision and expansive scale.

\noindent\bftext{b13.} The camera swiftly navigates through a labyrinthine boiler room, weaving through a dense network of pipes, valves, and gauges. It makes sharp, precise right-angle turns at each junction, capturing the intricate industrial landscape. The scene is bathed in steady, even lighting, highlighting the metallic sheen of the equipment. The walkways are lined with sturdy gratings, and the camera's movement is fluid and rapid, yet the environment remains eerily still, with no vibrations or steam emissions. The constant exposure maintains a clear view of the complex machinery, while the camera's agile dance creates a dynamic visual journey through the mechanical maze.

\noindent\bftext{b14.} The camera begins its graceful descent from a grand balcony, capturing the intricate details of the ornate railings and plush, empty seats below. As it swoops downward, the theater's rich red and gold color scheme becomes more pronounced, with rows of seats stretching out like a sea of anticipation. The camera glides past the elegant acoustic panels lining the walls, their design enhancing the theater's opulent atmosphere. The steady, warm glow of the overhead lights bathes the scene in a golden hue, highlighting the theater's grandeur. As the camera approaches the stage, the polished wooden floor of the apron gleams under the lights, inviting the viewer into the heart of this majestic, silent space.

\noindent\bftext{b15.} The camera swiftly ascends a sleek, modern spiral staircase in a spacious office atrium, capturing the polished metal handrails and glass balusters in sharp detail. As it spirals upward, the lens brushes past minimalist signage, reflecting the building's contemporary design. The environment remains static and pristine, with the staircase's geometric precision emphasized by the constant radius of the climb. The atrium's ambient lighting casts subtle reflections on the surfaces, enhancing the sense of motion. The camera's ascent is smooth and uninterrupted, offering a dynamic perspective of the architectural elegance and the stillness of the surrounding space.

\noindent\bftext{b16.} The camera embarks on a dynamic barrel-roll journey through a luminous glass atrium corridor, where sunlight streams through expansive windows, casting intricate patterns on the polished floor. As the camera gracefully rotates 180 degrees, the corridor's features remain steadfast: sleek planters brimming with verdant foliage, modern benches inviting rest, and sleek directory boards offering guidance. The reflections on the glass walls and floor remain undisturbed, creating a mesmerizing kaleidoscope of light and shadow. The camera's fluid motion captures the essence of the space, transforming the atrium into a captivating dance of architecture and light.

\noindent\bftext{b17.} The camera swiftly navigates through a labyrinth of office cubicles, each enclosed by beige partitions, revealing a sea of identical workspaces. Desktops and monitors, displaying static spreadsheets and graphs, line the desks, while cable trays snake overhead, adding to the structured chaos. The camera makes a sharp ninety-degree turn, narrowly avoiding a stack of neatly organized files, before continuing its rapid journey. Another tight turn reveals more cubicles, each identical yet distinct, with personal touches like family photos and coffee mugs. The static screens and still environment contrast with the camera's relentless speed, creating a dynamic visual experience.

\noindent\bftext{b18.} The camera swiftly navigates a library's cross-aisle, darting between towering bookshelves densely packed with colorful spines and neatly labeled endcaps. As it races forward, the camera captures the intricate details of book titles and the orderly arrangement of carts and signs, all fixed in place. The lighting remains steady, casting a warm glow over the scene, enhancing the rich textures of the books and the polished wood of the shelves. Suddenly, the camera pivots sharply around a freestanding shelf, revealing a new perspective of the library's labyrinthine layout, before continuing its rapid journey through the quiet, knowledge-filled sanctuary.

\noindent\bftext{b19.} The camera swiftly glides through a bustling electronics store aisle, capturing rows of neatly stacked boxed products and sleek demo stations. The shelves are lined with vibrant packaging, showcasing the latest gadgets and devices. As the camera moves, it reveals a variety of electronics, from headphones to smart speakers, all meticulously arranged. The store is filled with the hum of quiet chatter and the soft rustle of packaging. The camera's journey culminates in a dramatic halt at a central display, where a large, prominent screen stands, its surface reflecting the ambient light. Despite the dynamic sweep, all screens remain static, their blank faces contrasting with the lively atmosphere, emphasizing the anticipation of technology yet to be activated.

\noindent\bftext{b20.} The camera swiftly descends through a spacious, modern elevator lobby atrium, capturing the sleek design of the space. Rows of metallic mailboxes line the walls, their polished surfaces reflecting the ambient light. Elegant signage, with clear, bold lettering, guides visitors through the area. As the camera glides downward, it passes over a series of sleek, stainless steel turnstiles, their surfaces gleaming under the atrium's soft lighting. The descent continues from the mezzanine rail height, offering a panoramic view of the lobby's architectural details, down to the polished marble floor. The camera advances steadily toward the closed elevator doors, which stand as a focal point. Throughout the descent, the elevator indicators remain steady, their lights unwavering, emphasizing the smooth, uninterrupted motion of the camera's journey.

\subsection{Outdoor Prompts}
\subsubsection{Slow camera movements}

\noindent\bftext{c1.} The camera glides slowly through a serene cloister courtyard, enveloped by elegant arcades on each side. The stone columns, adorned with intricately carved capitals, stand as silent sentinels beneath the roofline, their artistry captured in exquisite detail. The courtyard's floor is a mosaic of aged stone tiles, each telling its own story. The ambient light bathes the scene in a soft, timeless glow, casting gentle shadows that dance across the surfaces. As the camera advances, the tranquility of the space is palpable, with the arches framing views of the lush, manicured garden at the center, inviting contemplation and peace.

\noindent\bftext{c2.} In a narrow, ancient stone alley, the camera glides smoothly, capturing the intricate textures of weathered facades, where every crack and crevice tells a story of time. The alley is flanked by towering stone walls, their surfaces adorned with ornate reliefs and timeworn lintels, each doorway a portal to history. The atmosphere is still and silent, with no breeze to disturb the rigid tranquility. As the camera orbits, it remains close to the walls, emphasizing the craftsmanship and detail of the stonework, while the alley's narrow confines create an intimate, almost claustrophobic ambiance, leaving the distant world unseen and mysterious.

\noindent\bftext{c3.} The camera smoothly glides along a row of vibrant shipping containers, stacked three high, their bold colors contrasting against the muted sky. The containers, in shades of red, blue, and green, stand in perfect alignment, their surfaces weathered by time and travel. The yard is enclosed by a sturdy chain-link fence, its metallic sheen catching the light, while the tops of the containers remain just below the frame's upper edge, creating a sense of containment and order. The scene is still and silent, with no movement or fluttering fabric, emphasizing the solitude and industrial beauty of the setting as the camera tracks steadily along the row.

\noindent\bftext{c4.} The camera glides smoothly beneath a series of massive concrete bridge arches, capturing the intricate details of the structure. Each arch, with its weathered surface and subtle variations in texture, stands as a testament to engineering prowess. The camera moves gracefully from one pier to the next, maintaining a steady path under the deck, revealing the symmetry and strength of the construction. The surrounding environment is serene, with the gentle sound of water echoing softly. Light filters through the gaps, casting dynamic shadows that dance across the concrete, enhancing the sense of depth and perspective in this tranquil, architectural journey.

\noindent\bftext{c5.} The camera smoothly glides around a quaint, enclosed plaza, its cobblestone floor echoing the whispers of history. At the heart of this serene courtyard stands a majestic stone statue, its features meticulously carved, exuding an aura of timeless grace. The surrounding walls, aged and weathered, form a protective embrace, their surfaces adorned with creeping ivy and subtle cracks that tell tales of the past. The light remains constant, casting gentle shadows that dance across the stone, enhancing the statue's dignified presence. As the camera circles, the scene remains tranquil, capturing the essence of stillness and solitude within this hidden sanctuary.

\noindent\bftext{c6.} The camera glides smoothly across a sunlit brick courtyard, enclosed by towering, weathered walls, each brick displaying unique textures and subtle color variations. The lens captures the intricate details of the aged brickwork, revealing the craftsmanship of the past. As the camera moves, it focuses on the ornate window frames, their wooden surfaces worn by time, and the sturdy, iron-clad door frames, each telling a silent story of resilience. The fixed exposure and focus highlight the play of light and shadow across the surfaces, creating a serene, timeless atmosphere. The absence of movement in the environment emphasizes the stillness, allowing the viewer to appreciate the architectural details within the ten-meter range.

\noindent\bftext{c7.} The camera glides smoothly through a serene, enclosed garden, where towering hedges form lush, green walls, their leaves and branches perfectly still, creating a sense of tranquility. Stone paths weave intricately beneath the camera's gentle orbit, revealing a mosaic of cobblestones and pebbles, each step echoing the garden's timeless elegance. The camera remains low, capturing the rich textures of the hedges, their leaves a vibrant tapestry of greens, while the absence of the horizon enhances the garden's secluded charm. Sunlight filters softly through the foliage, casting delicate patterns on the paths, as the camera continues its graceful, uninterrupted circle.

\noindent\bftext{c8.} A steady camera glides through a covered market arcade, capturing the quiet ambiance of closed stalls and rigid awnings. The path is lined with colorful, yet motionless, signs and awnings, each displaying faded logos and names, hinting at the market's bustling past. The camera's smooth movement contrasts with the stillness of the scene, emphasizing the absence of people and activity. The arcade's architecture, with its intricate ironwork and vintage lamps, adds a nostalgic charm. As the camera advances, the muted colors and soft lighting create a serene, almost timeless atmosphere, evoking a sense of calm and reflection.

\noindent\bftext{c9.} The camera glides smoothly along a meticulously constructed scaffolding, tightly enveloping a historic masonry facade. The structure's intricate network of vertical poles and horizontal planks creates a geometric maze, inviting the viewer to explore its depths. As the camera weaves through this industrial labyrinth, the surrounding tarps hang still, their surfaces adorned with colorful tags and markings, each telling a silent story of the ongoing restoration. The lighting remains unwavering, casting a soft, even glow that highlights the texture of the weathered stone and the scaffolding's metallic sheen. The only movement is the camera's deliberate journey, capturing the serene stillness of this architectural cocoon.

\noindent\bftext{c10.} A deliberate dolly shot captures the side of a parked freight locomotive, its vibrant colors contrasting against the muted tones of a towering retaining wall. The camera glides smoothly alongside the train, revealing intricate details of the locomotive's exterior, from its weathered metal surface to the bold insignia emblazoned on its side. The background remains consistently framed by the high retaining wall and a row of parked cars, emphasizing the train's imposing presence. The scene is serene, with no movement from hoses or cables, allowing the viewer to focus solely on the locomotive's grandeur as the camera continues its steady slide.

\subsubsection{Fast camera movements}

\noindent\bftext{c11.} The camera embarks on a swift journey through a scaffold tunnel, enclosing a bustling sidewalk, capturing the essence of urban life. The structure, devoid of tarps or fluttering tags, stands rigid and industrial, its metallic framework casting intricate shadows on the ground. As the camera races through, it weaves skillfully between the vertical beams, maintaining a seamless flow within the scaffold's confines. The tunnel's linear perspective creates a mesmerizing visual rhythm, with the steady, unyielding lines of the scaffold contrasting against the dynamic motion of the camera, evoking a sense of speed and precision in this urban passageway.

\noindent\bftext{c12.} The camera swiftly navigates through a dimly lit parking structure, weaving through ground-level lanes bordered by imposing concrete walls and sturdy columns. The scene is eerily still, with parked cars, directional signs, and various fixtures frozen in time, creating a stark contrast to the camera's dynamic movement. As it executes two precise, sharp turns, the camera captures the texture of the concrete and the subtle play of shadows, enhancing the sense of speed and agility. The atmosphere is tense yet captivating, with the camera's fluid motion providing a sense of urgency and exploration within the static environment.

\noindent\bftext{c13.} In a secluded, ancient stone courtyard, a majestic fountain stands at its center, crafted from weathered marble, its intricate carvings telling tales of old. The camera swiftly arcs around the fountain, capturing every detail of its ornate design, while the surrounding high stone walls, adorned with creeping ivy, create an intimate, enclosed atmosphere. The water within the fountain remains perfectly still, a flawless mirror reflecting the fountain's grandeur and the muted tones of the stone. The rapid camera movement contrasts with the serene, undisturbed water, emphasizing the tranquility and timelessness of this hidden sanctuary.

\noindent\bftext{c14.} The camera swiftly navigates through a deserted market arcade, gliding under a series of closed stalls and vibrant awnings, each adorned with colorful, fixed signs. The lighting remains consistently warm, casting a gentle glow over the scene, enhancing the rich textures of the fabric and wood. As the camera accelerates, it executes a graceful S-curve, weaving smoothly between the stalls, capturing the intricate details of the market's architecture. The fixed awnings flutter slightly in the breeze, adding a sense of life to the otherwise still environment, while the constant speed of the camera creates a dynamic, immersive experience.

\noindent\bftext{c15.} The camera ascends swiftly up a narrow, dimly lit exterior stairwell, flanked by two towering brick facades, their textures detailed and weathered. The stairwell is a tight, vertical corridor, with each step meticulously crafted from aged stone, worn by time. The walls on either side are adorned with creeping ivy and faded graffiti, adding character to the urban setting. As the camera climbs, the ambient sounds of a bustling city are faintly audible, yet the stairwell remains eerily still, creating a sense of isolation. The ascent is smooth and uninterrupted, with the facades maintaining their imposing presence, emphasizing the confined space and the absence of any distant horizon.

\noindent\bftext{c16.} A dynamic aerial view captures a robust freight locomotive, its vibrant colors contrasting against the towering high retaining wall, as the camera swiftly orbits around it. The locomotive, with its intricate details and weathered exterior, stands motionless on the tracks, surrounded by an expansive, quiet rail yard. The camera's rapid circular motion creates a dizzying effect, emphasizing the locomotive's imposing presence and the wall's textured surface. As the orbit tightens, the locomotive's features become more pronounced, showcasing its powerful build and the industrial ambiance of the yard, while the retaining wall remains a constant, looming backdrop.

\noindent\bftext{c17.} The camera races through a narrow brick service alley, capturing the gritty details of the urban landscape. The walls, close and confining, are lined with weathered doors, rusted meters, and overflowing dumpsters, each telling a story of the city's hidden life. The bricks, aged and stained, create a textured backdrop as the camera speeds past, maintaining a tight focus to emphasize the claustrophobic nature of the alley. The scene is static, with no movement from the objects, enhancing the sensation of speed and urgency. The alley's dim lighting casts long shadows, adding to the mysterious and slightly ominous atmosphere, as the camera's swift journey continues uninterrupted.

\noindent\bftext{c18.} The camera swiftly circles a grand marble statue of a mythological figure, positioned at the heart of a small, serene museum courtyard. The courtyard is enclosed by towering, ivy-covered walls, creating an intimate, secluded atmosphere. As the camera loops rapidly, it brushes past intricately carved stone columns and ornate wrought-iron railings, capturing the play of light and shadow on their surfaces. The statue, bathed in soft, natural light filtering from above, stands as the focal point amidst the stillness, its detailed features and flowing robes frozen in time. The only movement is the camera's dynamic orbit, emphasizing the tranquil, timeless ambiance of the courtyard.

\noindent\bftext{c19.} The camera glides swiftly through an expansive underpass colonnade, capturing the rhythmic alignment of towering concrete columns. The scene is devoid of any movement from flags, foliage, or debris, emphasizing the stillness of the environment. The camera's fluid motion creates a seamless journey, weaving gracefully between the columns, maintaining a steady path beneath the deck. The play of light and shadow on the concrete surfaces adds depth and texture, enhancing the visual experience. The continuous movement offers a mesmerizing perspective, drawing the viewer into the architectural symmetry and the serene, undisturbed atmosphere of the underpass.

\noindent\bftext{c20.} The camera swiftly navigates through a labyrinth of towering shipping containers, their vibrant hues of red, blue, and green contrasting against the clear sky. The narrow lanes create a sense of urgency as the camera glides effortlessly, making two sharp ninety-degree turns, each corner revealing a new path. The static containers, adorned with international shipping logos and weathered signs, stand as silent sentinels, their metallic surfaces reflecting the sunlight. The camera's rapid movement through the maze creates a thrilling sense of exploration, capturing the intricate geometry and industrial beauty of the container cityscape.

\subsection{Appearance-Complex Prompts}
\subsubsection{Slow camera movements}

\noindent\bftext{d1.} A sleek chrome motorcycle stands proudly indoors, its polished surface reflecting the static environment. The camera begins a slow, deliberate orbit around the bike, capturing the intricate details of its curved bodywork. As the camera glides smoothly, stable reflections slide gracefully across the chrome, highlighting the craftsmanship and design. The environment remains rigid and still, with no movement or change in lighting, emphasizing the motorcycle's gleaming surface. The constant exposure and lighting create a serene atmosphere, allowing the viewer to appreciate the motorcycle's elegance and the artistry of its reflective surfaces.

\noindent\bftext{d2.} The camera glides smoothly along a wall of glass bricks, each brick capturing and distorting the static room beyond with mesmerizing refractions. The bricks, arranged in a precise grid, create a kaleidoscope effect, bending light and subtly altering the view of the room's muted tones and simple furnishings. As the camera maintains a constant distance, the glass bricks reveal a dance of light and shadow, with each brick acting as a lens, offering a unique perspective of the stillness beyond. The gentle dolly movement emphasizes the serene, unchanging nature of the room, while the glass bricks transform the ordinary into a captivating visual symphony.

\noindent\bftext{d3.} The camera glides smoothly across an intricate mosaic floor, capturing the mesmerizing dance of geometric patterns and vibrant colors. Each tile, a unique piece of art, is meticulously arranged, creating a harmonious tapestry of shapes and hues. The grout lines, precise and unwavering, form a delicate lattice that binds the mosaic together. The camera's shallow angle reveals the subtle texture of the tiles, highlighting their glossy surfaces and the occasional imperfections that add character. As the camera slides laterally, the fixed exposure and focus maintain a consistent clarity, allowing the viewer to appreciate the artistry and craftsmanship of the mosaic in exquisite detail.

\noindent\bftext{d4.} Inside a serene, dimly lit museum gallery, a camera smoothly advances toward an exquisite display of crystal sculptures encased in pristine glass. The sculptures, intricately carved with precision, stand motionless, their multifaceted surfaces capturing and reflecting the ambient light. As the camera draws closer, the stable specular highlights glide gracefully over the crystal facets, creating a mesmerizing dance of light and shadow. The glass case remains untouched, preserving the delicate artistry within. The steady push-in reveals the intricate details of each sculpture, emphasizing their brilliance and craftsmanship, while the ambient lighting remains constant, enhancing the serene and contemplative atmosphere.

\noindent\bftext{d5.} The camera glides slowly along a wire-mesh fence, capturing the intricate lattice of thin metal strands, each intersection forming a precise grid pattern. The focus shifts to the scaffolding joint, where metallic beams intersect, showcasing the industrial elegance of their design. The camera's steady movement highlights the repeating gaps, creating a rhythmic visual pattern. The scene is devoid of motion or distortion, with every element in perfect alignment, emphasizing the rigidity and precision of the structure. The lighting casts subtle shadows, enhancing the texture and depth of the metal surfaces, while the camera's unwavering track maintains a serene, mechanical grace.

\noindent\bftext{d6.} The camera glides smoothly over an expansive circuit board, capturing the intricate landscape of technology. It begins at the robust connectors, their metallic surfaces gleaming under soft lighting. As it moves, the lens reveals a network of fine copper traces, like delicate veins, weaving across the board's surface. The journey continues over an array of components: capacitors, resistors, and microchips, each precisely placed, their labels and markings crisp and clear. The board's surface is a sea of green, punctuated by the occasional silver and black of the components. The scene is serene, with no blinking lights or moving parts, emphasizing the stillness and precision of the electronic world.

\noindent\bftext{d7.} The camera glides slowly along a perforated metal screen wall, revealing a mesmerizing pattern of repeating circular holes. The oxidized texture of the metal, with its rich hues of rust and patina, creates a tapestry of earthy tones. As the camera maintains a constant angle, the intricate details of the screen's surface become apparent, showcasing the interplay of light and shadow across the perforations. The static surroundings, including a hint of industrial architecture in the background, remain unchanged, emphasizing the serene, rhythmic motion of the camera's lateral journey. The scene evokes a sense of timelessness and industrial beauty.

\noindent\bftext{d8.} The camera begins its slow, deliberate journey toward a meticulously crafted wicker basket resting on a rustic wooden table. The basket's intricate weave, a tapestry of interlocking fibers, becomes increasingly detailed as the camera draws nearer. The ambient lighting casts a warm, golden hue, accentuating the basket's natural tones and textures. The table's rich, grainy surface complements the basket's earthy aesthetic. As the camera continues its steady approach, the basket's craftsmanship is revealed in stunning clarity, each fiber distinct and unmoving, creating a serene, almost meditative visual experience. The focus remains sharp, capturing the essence of artisanal skill.

\noindent\bftext{d9.} A sleek, polished black grand piano stands majestically in a spacious, elegantly lit room, its glossy surface reflecting the ambient light. The camera begins a gentle orbit around the piano, capturing the smooth, stable reflections that glide effortlessly across the curved lid, creating a mesmerizing dance of light and shadow. The room is tastefully decorated, with soft, neutral tones that complement the piano's deep black finish. As the camera continues its steady circle, the intricate details of the piano's craftsmanship are revealed, from the delicate curve of its legs to the subtle sheen of its keys. The atmosphere is serene and timeless, with the constant exposure and white balance maintaining a harmonious visual experience, allowing the viewer to fully appreciate the piano's elegance and the tranquil beauty of the setting.

\noindent\bftext{d10.} The camera glides slowly across a set of meticulously aligned Venetian blinds, each slat casting sharp, parallel shadows that create a mesmerizing pattern of high-frequency lines. The blinds are illuminated from the left, casting a warm, consistent glow that highlights their rigid structure. The angle remains shallow, emphasizing the precision and uniformity of the slats, which stand perfectly still, devoid of any movement. The steady lighting accentuates the texture and depth of the blinds, while the camera's smooth lateral motion reveals the intricate play of light and shadow, creating a hypnotic visual rhythm.

\subsubsection{Fast camera movements}

\noindent\bftext{d11.} In a sleek, mirrored showroom, the camera embarks on a swift orbit, weaving seamlessly between four towering mirrors. The reflective surfaces, pristine and flawless, capture the dynamic motion, creating an illusion of infinite space. The room's polished marble floor and minimalist decor, featuring a single elegant vase on a pedestal, remain perfectly still, enhancing the sense of motion. As the camera glides rapidly, the mirrors reflect each other, forming a mesmerizing kaleidoscope effect. The ambient lighting casts a soft, even glow, ensuring no flicker disrupts the visual harmony. The rapid movement creates a captivating dance of reflections, maintaining focus within the room's boundaries.

\noindent\bftext{d12.} The camera swiftly navigates through narrow aisles of a high-tech server room, where matte black panels line the path, each adorned with steady, glowing indicator lights in hues of green and blue. The atmosphere is silent and still, with the only movement being the camera's agile journey. It executes two precise, sharp turns, maintaining a close proximity to the towering racks, emphasizing the dense, labyrinthine layout. The ambient hum of the servers is palpable, yet nothing blinks or shifts, creating a sense of calm amidst the technological maze, as the camera's fluid motion contrasts with the static environment.

\noindent\bftext{d13.} A sleek, high-definition lateral sweep across a set of Venetian blinds, each slat meticulously aligned, captures the essence of motion and light. The blinds, illuminated from the left, cast sharp, rhythmic shadows, creating a mesmerizing pattern of light and dark. As the camera glides smoothly from one end to the other, the steady illumination highlights the precision of each slat, emphasizing the aliasing effect. The blinds remain perfectly still, their metallic sheen reflecting the consistent light source, while the camera's swift, unwavering movement creates a dynamic visual experience, showcasing the interplay of light, shadow, and motion.

\noindent\bftext{d14.} The camera embarks on a dynamic journey through a complex network of aluminum pipes, suspended intricately beneath a vast factory ceiling. The pipes, gleaming under industrial lights, form a maze of intersecting pathways at diverse heights, creating a labyrinthine structure. As the camera weaves swiftly through the rigid, motionless framework, it navigates tight corners and narrow passages, offering a thrilling perspective of the metallic landscape. The pipes' polished surfaces reflect the ambient light, casting intricate shadows on the factory floor below, enhancing the sense of depth and complexity in this industrial marvel.

\noindent\bftext{d15.} The camera swiftly tilts and dollies around a grand glass display case, filled with an array of dazzling cut gemstones, each facet catching the light with precision. As the camera circles one corner, the gems remain perfectly still, their vibrant colors and intricate cuts highlighted by consistent specular reflections. The brisk movement of the camera creates a dynamic perspective, emphasizing the brilliance and clarity of the gemstones. Exiting to a wider view, the display case stands majestically in the center of an elegant showroom, its contents shimmering under the soft, ambient lighting, inviting admiration and awe.

\noindent\bftext{d16.} The camera ascends a meticulously crafted spiral staircase within a lattice tower, composed of slender, interwoven steel members. The structure's geometry is precise, maintaining a consistent radius as it spirals upward, creating a mesmerizing visual rhythm. The steel members are rigid and unwavering, their linear forms intersecting at calculated angles, forming a harmonious lattice pattern. The lighting is steady, casting soft, even illumination across the metallic surfaces, highlighting their sleek, industrial elegance. As the camera climbs, the viewer experiences a seamless ascent, the tower's intricate design unfolding in a continuous, hypnotic motion, evoking a sense of infinite elevation.

\noindent\bftext{d17.} The camera swiftly glides along a gallery wall, capturing a dynamic sequence of framed fine-text posters and signage, each meticulously aligned. The frames, rigid and unyielding, display intricate typography and vivid graphics, their details momentarily blurred by the rapid motion. The shallow grazing angle accentuates the depth and texture of the wall, creating a sense of urgency and fluidity. As the camera races past, the posters remain unchanged, their content a constant amidst the motion. The sequence culminates in a precise halt, centering on a prominent frame, its text and imagery now in sharp focus, inviting contemplation and appreciation.

\noindent\bftext{d18.} The camera swiftly navigates through a labyrinth of hanging beaded curtains, each strand composed of shimmering glass spheres that catch and reflect ambient light in a mesmerizing dance of colors. As the camera weaves between two parallel strands, the beads remain perfectly still, their reflections creating a kaleidoscope of stable, prismatic patterns. The journey continues with a fluid motion, the camera gliding effortlessly past the strands, capturing the intricate details of each glass sphere. Finally, the camera exits gracefully past a sleek display plinth, leaving behind the serene, crystalline world of suspended glass, where only the camera's movement disturbs the tranquil beauty.

\noindent\bftext{d19.} The camera glides smoothly through an expansive hall, where towering tile columns and elegant archways repeat in perfect symmetry, creating a mesmerizing visual rhythm. The tiles, intricately patterned in shades of blue and white, reflect the ambient light, casting subtle shadows that dance across the floor. As the camera performs an S-curve, it remains nestled beneath the grand vaults, capturing the majestic scale and architectural precision of the space. The rigid columns and arches stand immobile, their timeless beauty accentuated by the fluid motion of the camera, which weaves gracefully through the serene, echoing corridor.

\noindent\bftext{d20.} The camera glides smoothly through a serene sculpture gallery, where glossy black ceramic pieces rest elegantly on sleek pedestals. Each sculpture, meticulously crafted, reflects ambient light, creating a mesmerizing play of shadows and highlights. The camera weaves gracefully between the artworks, capturing the intricate details and the polished surfaces that mirror their surroundings. As it approaches the gallery's centerpiece, a strikingly intricate sculpture, the camera executes a tight, fluid turn, offering a dynamic perspective of the central piece. Throughout the journey, the sculptures remain perfectly still, their reflections unwavering, while the camera's motion brings the gallery to life.


%
%
\bibliographystyle{splncs04}
\bibliography{main}

@String(CVPR  = {IEEE Conf. Comput. Vis. Pattern Recog.})

@String(ICCV  = {Int. Conf. Comput. Vis.})

@String(ECCV  = {Eur. Conf. Comput. Vis.})

@String(NeurIPS = {Adv. Neural Inform. Process. Syst.})

@String(ICML  = {Int. Conf. Mach. Learn.})

@String(ICLR  = {Int. Conf. Learn. Represent.})

@String(BMVC  = {Brit. Mach. Vis. Conf.})

@String(CVPRW = {IEEE Conf. Comput. Vis. Pattern Recog. Worksh.})

@String(CVPR  = {CVPR})

@String(ICCV  = {ICCV})

@String(ECCV  = {ECCV})

@String(NeurIPS = {NeurIPS})

@String(ICML  = {ICML})

@String(ICLR  = {ICLR})

@String(BMVC  =	{BMVC})

@String(CVPRW = {CVPRW})

@inproceedings{Asim2025MEt3RMM,
  title={{MEt3R}: Measuring Multi-View Consistency in Generated Images},
  author={Mohammad Asim and Christopher Wewer and Thomas Wimmer and Bernt Schiele and Jan Eric Lenssen},
  booktitle=CVPR,
  year={2025},
  pages={6034-6044},
}

@article{wan2025wan,
  title={Wan: Open and advanced large-scale video generative models},
  author={Wan, Team and Wang, Ang and Ai, Baole and Wen, Bin and Mao, Chaojie and Xie, Chen-Wei and Chen, Di and Yu, Feiwu and Zhao, Haiming and Yang, Jianxiao and others},
  journal={arXiv preprint arXiv:2503.20314},
  year={2025}
}

@article{yang2024cogvideox,
  title={{CogVideoX}: Text-to-Video Diffusion Models with An Expert Transformer},
  author={Yang, Zhuoyi and Teng, Jiayan and Zheng, Wendi and Ding, Ming and Huang, Shiyu and Xu, Jiazheng and Yang, Yuanming and Hong, Wenyi and Zhang, Xiaohan and Feng, Guanyu and others},
  journal={arXiv preprint arXiv:2408.06072},
  year={2024}
}

@article{hong2022cogvideo,
  title={{CogVideo}: Large-scale Pretraining for Text-to-Video Generation via Transformers},
  author={Hong, Wenyi and Ding, Ming and Zheng, Wendi and Liu, Xinghan and Tang, Jie},
  journal={arXiv preprint arXiv:2205.15868},
  year={2022}
}

@inproceedings{zhang2025ufm,
 title={{UFM}: A Simple Path towards Unified Dense Correspondence with Flow},
 author={Zhang, Yuchen and Keetha, Nikhil and Lyu, Chenwei and Jhamb, Bhuvan and Chen, Yutian and Qiu, Yuheng and Karhade, Jay and Jha, Shreyas and Hu, Yaoyu and Ramanan, Deva and Scherer, Sebastian and Wang, Wenshan},
 booktitle={arXiV},
 year={2025}
}

@inproceedings{Yu2023PhotoconsistentNVS,
  title={Long-Term Photometric Consistent Novel View Synthesis with Diffusion Models},
  author={Jason J. Yu and Fereshteh Forghani and Konstantinos G. Derpanis and Marcus A. Brubaker},
  booktitle=ICCV,
  year={2023},
}

@inproceedings{reizenstein21co3d,
	Author = {Reizenstein, Jeremy and Shapovalov, Roman and Henzler, Philipp and Sbordone, Luca and Labatut, Patrick and Novotny, David},
	Booktitle = ICCV,
	Title = {Common Objects in {{3D}}: Large-Scale Learning and Evaluation of Real-life 3D Category Reconstruction},
	Year = {2021},
}

@inproceedings{duan2025worldscore,
  title={World{S}core: A Unified Evaluation Benchmark for World Generation},
  author={Duan, Haoyi and Yu, Hong-Xing and Chen, Sirui and Fei-Fei, Li and Wu, Jiajun},
  booktitle=ICCV,
  year={2025}
}

@inproceedings{dai2017scannet,
    title={{ScanNet}: Richly-annotated 3{D} Reconstructions of Indoor Scenes},
    author={Dai, Angela and Chang, Angel X. and Savva, Manolis and Halber, Maciej and Funkhouser, Thomas and Nie{\ss}ner, Matthias},
    booktitle = CVPR,
    year = {2017}
}

@inproceedings{yeshwanthliu2023scannetpp,
  title={{ScanNet}++: A High-Fidelity Dataset of 3{D} Indoor Scenes},
  author={Yeshwanth, Chandan and Liu, Yueh-Cheng and Nie{\ss}ner, Matthias and Dai, Angela},
  booktitle = ICCV,
  year={2023}
}

@inproceedings{wang2025vggt,
  title={{VGGT}: Visual Geometry Grounded Transformer},
  author={Wang, Jianyuan and Chen, Minghao and Karaev, Nikita and Vedaldi, Andrea and Rupprecht, Christian and Novotny, David},
  booktitle=CVPR,
  year={2025}
}

@article{Jang2025FrameGT,
  title={Frame Guidance: Training-Free Guidance for Frame-Level Control in Video Diffusion Models},
  author={Sang-Sub Jang and Taekyung Ki and Jaehyeong Jo and Jaehong Yoon and Soo Ye Kim and Zhe L. Lin and Sung Ju Hwang},
  journal={ArXiv},
  year={2025},
  volume={abs/2506.07177}
}

@inproceedings{zhang2023adding,
  title={Adding conditional control to text-to-image diffusion models},
  author={Zhang, Lvmin and Rao, Anyi and Agrawala, Maneesh},
  booktitle=ICCV,
  pages={3836--3847},
  year={2023}
}

@inproceedings{
    zhang2305controlvideo,
    title={{ControlVideo}: Training-free Controllable Text-to-video Generation},
    author={Yabo Zhang and Yuxiang Wei and Dongsheng Jiang and XIAOPENG ZHANG and Wangmeng Zuo and Qi Tian},
    booktitle=ICLR,
    year={2024},
}

@inproceedings{Bansal2023UniversalGF,
  title={Universal Guidance for Diffusion Models},
  author={Arpit Bansal and Hong-Min Chu and Avi Schwarzschild and Soumyadip Sengupta and Micah Goldblum and Jonas Geiping and Tom Goldstein},
  booktitle={CVPRW},
  year={2023},
  pages={843-852},
}

@article{Nam2024OpticalFlowGP,
  title={Optical-Flow Guided Prompt Optimization for Coherent Video Generation},
  author={Hyelin Nam and Jaemin Kim and Dohun Lee and Jong Chul Ye},
  journal=CVPR,
  year={2024},
  pages={7837-7846},
}

@inproceedings{geng2024motion,
  author    = {Geng, Daniel and Owens, Andrew},
  title     = {Motion Guidance: Diffusion-Based Image Editing with Differentiable Motion Estimators},
  booktitle = ICLR,
  year      = {2024},
}

@article{Huang2024VBenchCA,
  title={{VBench}++: Comprehensive and Versatile Benchmark Suite for Video Generative Models},
  author={Ziqi Huang and Fan Zhang and Xiaojie Xu and Yinan He and Jiashuo Yu and Ziyue Dong and Qianli Ma and Nattapol Chanpaisit and Chenyang Si and Yuming Jiang and Yaohui Wang and Xinyuan Chen and Yingcong Chen and Limin Wang and Dahua Lin and Yu Qiao and Ziwei Liu},
  journal={ArXiv},
  year={2024},
}

@article{Huang2023VBenchCB,
  title={{VBench}: Comprehensive Benchmark Suite for Video Generative Models},
  author={Ziqi Huang and Yinan He and Jiashuo Yu and Fan Zhang and Chenyang Si and Yuming Jiang and Yuanhan Zhang and Tianxing Wu and Qingyang Jin and Nattapol Chanpaisit and Yaohui Wang and Xinyuan Chen and Limin Wang and Dahua Lin and Yu Qiao and Ziwei Liu},
  journal=CVPR,
  year={2023},
  pages={21807-21818},
}

@article{Liu2023EvalCrafterBA,
  title={{EvalCrafter}: Benchmarking and Evaluating Large Video Generation Models},
  author={Yaofang Liu and Xiaodong Cun and Xuebo Liu and Xintao Wang and Yong Zhang and Haoxin Chen and Yang Liu and Tieyong Zeng and Raymond H. Chan and Ying Shan},
  journal=CVPR,
  year={2023},
  pages={22139-22149},
}

@article{han2024multistable,
  title={Multistable shape from shading emerges from patch diffusion},
  author={Han, Xinran and Zickler, Todd and Nishino, Ko},
  journal=NeurIPS,
  volume={37},
  pages={34686--34711},
  year={2024}
}

@inproceedings{huang2025jog3r,
  title={{JOG3R}: Towards 3{D}-Consistent Video Generators},
  author={Huang, Chun-Hao Paul and Mitra, Niloy and Jeong, Hyeonho and Yoon, Jae Shin and Ceylan, Duygu},
  booktitle=BMVC,
  year={2025}
}

@inproceedings{ren2025gen3c,
  title={Gen3{C}: 3{D}-informed world-consistent video generation with precise camera control},
  author={Ren, Xuanchi and Shen, Tianchang and Huang, Jiahui and Ling, Huan and Lu, Yifan and Nimier-David, Merlin and M{\"u}ller, Thomas and Keller, Alexander and Fidler, Sanja and Gao, Jun},
  booktitle=CVPR,
  pages={6121--6132},
  year={2025}
}

@inproceedings{zhang2025world,
  title={World-consistent video diffusion with explicit 3{D} modeling},
  author={Zhang, Qihang and Zhai, Shuangfei and Martin, Miguel Angel Bautista and Miao, Kevin and Toshev, Alexander and Susskind, Joshua and Gu, Jiatao},
  booktitle=CVPR,
  pages={21685--21695},
  year={2025}
}

@inproceedings{wang2024dust3r,
  title={{DUSt3R}: Geometric 3{D} vision made easy},
  author={Wang, Shuzhe and Leroy, Vincent and Cabon, Yohann and Chidlovskii, Boris and Revaud, Jerome},
  booktitle=CVPR,
  pages={20697--20709},
  year={2024}
}

@article{bhat2023zoedepth,
  title={Zoe{D}epth: Zero-shot transfer by combining relative and metric depth},
  author={Bhat, Shariq Farooq and Birkl, Reiner and Wofk, Diana and Wonka, Peter and M{\"u}ller, Matthias},
  journal={arXiv preprint arXiv:2302.12288},
  year={2023}
}

@article{yang2024depth,
  title={Depth anything {V}2},
  author={Yang, Lihe and Kang, Bingyi and Huang, Zilong and Zhao, Zhen and Xu, Xiaogang and Feng, Jiashi and Zhao, Hengshuang},
  journal=NeurIPS,
  volume={37},
  pages={21875--21911},
  year={2024}
}

@article{teed2021droid,
  title={{DROID-SLAM}: Deep visual slam for monocular, stereo, and {RGB-D} cameras},
  author={Teed, Zachary and Deng, Jia},
  journal=NeurIPS,
  volume={34},
  pages={16558--16569},
  year={2021}
}

@article{Lugmayr2022RePaintIU,
  title={{RePaint}: Inpainting using Denoising Diffusion Probabilistic Models},
  author={Andreas Lugmayr and Martin Danelljan and Andr{\'e}s Romero and Fisher Yu and Radu Timofte and Luc Van Gool},
  journal=CVPR,
  year={2022},
  pages={11451-11461}
}

@inproceedings{Wang2022ZeroShotIR,
  title={Zero-Shot Image Restoration Using Denoising Diffusion Null-Space Model},
  author={Yinhuai Wang and Jiwen Yu and Jian Zhang},
  booktitle=ICLR,
  year={2023}
}

@inproceedings{He2023ManifoldPG,
  title={Manifold Preserving Guided Diffusion},
  author={Yutong He and Naoki Murata and Chieh-Hsin Lai and Yuhta Takida and Toshimitsu Uesaka and Dongjun Kim and Wei-Hsiang Liao and Yuki Mitsufuji and J. Zico Kolter and Ruslan Salakhutdinov and Stefano Ermon},
  booktitle=ICLR,
  year={2024},
}

@inproceedings{huang2025segment,
  title={Segment Any Motion in Videos},
  author={Huang, Nan and Zheng, Wenzhao and Xu, Chenfeng and Keutzer, Kurt and Zhang, Shanghang and Kanazawa, Angjoo and Wang, Qianqian},
  booktitle=CVPR,
  pages={3406--3416},
  year={2025}
}

@article{kong2024hunyuanvideo,
  title={{HunyuanVideo}: A systematic framework for large video generative models},
  author={Kong, Weijie and Tian, Qi and Zhang, Zijian and Min, Rox and Dai, Zuozhuo and Zhou, Jin and Xiong, Jiangfeng and Li, Xin and Wu, Bo and Zhang, Jianwei and others},
  journal={arXiv preprint arXiv:2412.03603},
  year={2024}
}

@article{HaCohen2024LTXVideo,
  title={{LTX-Video}: Realtime Video Latent Diffusion},
  author={HaCohen, Yoav and Chiprut, Nisan and Brazowski, Benny and Shalem, Daniel and Moshe, Dudu and Richardson, Eitan and Levin, Eran and Shiran, Guy and Zabari, Nir and Gordon, Ori and Panet, Poriya and Weissbuch, Sapir and Kulikov, Victor and Bitterman, Yaki and Melumian, Zeev and Bibi, Ofir},
  journal={arXiv preprint arXiv:2501.00103},
  year={2024}
}

@misc{brooks2024sora,
  author       = {Tim Brooks and Bill Peebles and Connor Holmes and Will DePue
                  and Yufei Guo and Li Jing and David Schnurr and Joe Taylor
                  and Troy Luhman and Eric Luhman and Clarence Ng and Ricky Wang
                  and Aditya Ramesh},
  title        = {Video Generation Models as World Simulators},
  year         = {2024},
  note         = {{OpenAI technical report}}
}

@techreport{veo3techreport,
  title        = {Veo: a text-to-video generation system},
  author       = {{Google DeepMind}},
  year         = {2025},
  note         = {{Veo} 3 technical report},
}

@article{ravi2024sam2,
  title={{SAM} 2: Segment Anything in Images and Videos},
  author={Ravi, Nikhila and Gabeur, Valentin and Hu, Yuan-Ting and Hu, Ronghang and Ryali, Chaitanya and Ma, Tengyu and Khedr, Haitham and R{\"a}dle, Roman and Rolland, Chloe and Gustafson, Laura and Mintun, Eric and Pan, Junting and Alwala, Kalyan Vasudev and Carion, Nicolas and Wu, Chao-Yuan and Girshick, Ross and Doll{\'a}r, Piotr and Feichtenhofer, Christoph},
  journal={arXiv preprint arXiv:2408.00714},
  year={2024}
}

@article{agarwal2025cosmos,
  title={Cosmos world foundation model platform for physical ai},
  author={Agarwal, Niket and Ali, Arslan and Bala, Maciej and Balaji, Yogesh and Barker, Erik and Cai, Tiffany and Chattopadhyay, Prithvijit and Chen, Yongxin and Cui, Yin and Ding, Yifan and others},
  journal={arXiv preprint arXiv:2501.03575},
  year={2025}
}

@inproceedings{Lowe1999ObjectRF,
  title={Object recognition from local scale-invariant features},
  author={David G. Lowe},
  booktitle=ICCV,
  year={1999},
  volume={2},
  pages={1150-1157 vol.2},
}

@book{HartleyZisserman2003MVG,
  author    = {Hartley, Richard and Zisserman, Andrew},
  title     = {Multiple View Geometry in Computer Vision},
  publisher = {Cambridge University Press},
  year      = {2003}
}

@article{DeTone2017SuperPointSI,
  title={{SuperPoint}: Self-Supervised Interest Point Detection and Description},
  author={Daniel DeTone and Tomasz Malisiewicz and Andrew Rabinovich},
  journal={CVPRW},
  year={2017},
  pages={337-33712}
}

@inproceedings{Lindenberger2023LightGlueLF,
  title={{LightGlue}: Local Feature Matching at Light Speed},
  author={Philipp Lindenberger and Paul-Edouard Sarlin and Marc Pollefeys},
  booktitle=ICCV,
  year={2023},
  pages={17581-17592}
}

@article{Sampson1982FittingCS,
  title={Fitting conic sections to ``very scattered'' data: An iterative refinement of the bookstein algorithm},
  author={Paul D. Sampson},
  journal={Computer graphics and image processing},
  year={1982}
}

@inproceedings{Caron2021EmergingPI,
  title={Emerging Properties in Self-Supervised Vision Transformers},
  author={Mathilde Caron and Hugo Touvron and Ishan Misra and Herv'e J'egou and Julien Mairal and Piotr Bojanowski and Armand Joulin},
  booktitle=ICCV,
  year={2021},
  pages={9630-9640}
}

@inproceedings{Radford2021LearningTV,
  title={Learning Transferable Visual Models From Natural Language Supervision},
  author={Alec Radford and Jong Wook Kim and Chris Hallacy and Aditya Ramesh and Gabriel Goh and Sandhini Agarwal and Girish Sastry and Amanda Askell and Pamela Mishkin and Jack Clark and Gretchen Krueger and Ilya Sutskever},
  booktitle=ICML,
  year={2021}
}

@inproceedings{han2024vfusion3d,
  title={{VFusion3D}: Learning Scalable 3{D} Generative Models from Video Diffusion Models},
  author={Junlin Han and Filippos Kokkinos and Philip Torr},
  booktitle=ECCV,
  year={2024}
}

@inproceedings{voleti2024sv3d,
  title={{SV3D}: Novel multi-view synthesis and 3{D} generation from a single image using latent video diffusion},
  author={Voleti, Vikram and Yao, Chun-Han and Boss, Mark and Letts, Adam and Pankratz, David and Tochilkin, Dmitry and Laforte, Christian and Rombach, Robin and Jampani, Varun},
  booktitle=ECCV,
  pages={439--457},
  year={2024},
  organization={Springer}
}

@INPROCEEDINGS{VideoScene,
  author={Wang, Hanyang and Liu, Fangfu and Chi, Jiawei and Duan, Yueqi},
  booktitle=CVPR, 
  title={{VideoScene}: Distilling Video Diffusion Model to Generate 3D Scenes in One Step}, 
  year={2025},
  volume={},
  number={},
  pages={16475-16485},}

@article{zhang2025matrixgame,
  title     = {{Matrix-Game}: Interactive World Foundation Model},
  author    = {Yifan Zhang and Chunli Peng and Boyang Wang and Puyi Wang and Qingcheng Zhu and Fei Kang and Biao Jiang and Zedong Gao and Eric Li and Yang Liu and Yahui Zhou},
  journal   = {arXiv preprint arXiv:2506.18701},
  year      = {2025}
}

@article{he2025matrix,
    title={{Matrix-Game 2.0}: An Open-Source, Real-Time, and Streaming Interactive World Model},
    author={He, Xianglong and Peng, Chunli and Liu, Zexiang and Wang, Boyang and Zhang, Yifan and Cui, Qi and Kang, Fei and Jiang, Biao and An, Mengyin and Ren, Yangyang and Xu, Baixin and Guo, Hao-Xiang and Gong, Kaixiong and Wu, Cyrus and Li, Wei and Song, Xuchen and Liu, Yang and Li, Eric and Zhou, Yahui},
    journal={arXiv preprint arXiv:2508.13009},
    year={2025}
  }

@article{hyworld2025,
  title={{HY-World 1.5}: A Systematic Framework for Interactive World Modeling with Real-Time Latency and Geometric Consistency},
  author={Team HunyuanWorld},
  journal={arXiv preprint},
  year={2025}
}

@article{worldplay2025,
    title={WorldPlay: Towards Long-Term Geometric Consistency for Real-Time Interactive World Model},
    author={Wenqiang Sun and Haiyu Zhang and Haoyuan Wang and Junta Wu and Zehan Wang and Zhenwei Wang and Yunhong Wang and Jun Zhang and Tengfei Wang and Chunchao Guo},
    year={2025},
    journal={arXiv preprint}
}

@article{mao2025yume,
  title={Yume-1.5: A Text-Controlled Interactive World Generation Model},
  author={Mao, Xiaofeng and Li, Zhen and Li, Chuanhao and Xu, Xiaojie and Ying, Kaining and He, Tong and Pang, Jiangmiao and Qiao, Yu and Zhang, Kaipeng},
  journal={arXiv preprint arXiv:2512.22096},
  year={2025}
}

@misc{RelicWorldModel2025,
      title={{RELIC}: Interactive Video World Model with Long-Horizon Memory}, 
      author={Yicong Hong and Yiqun Mei and Chongjian Ge and Yiran Xu and Yang Zhou and Sai Bi and Yannick Hold-Geoffroy and Mike Roberts and Matthew Fisher and Eli Shechtman and Kalyan Sunkavalli and Feng Liu and Zhengqi Li and Hao Tan},
      year={2025},
      eprint={2512.04040},
      archivePrefix={arXiv},
      primaryClass={cs.CV}
}

@inproceedings{liu2024evalcrafter,
  title={Evalcrafter: Benchmarking and evaluating large video generation models},
  author={Liu, Yaofang and Cun, Xiaodong and Liu, Xuebo and Wang, Xintao and Zhang, Yong and Chen, Haoxin and Liu, Yang and Zeng, Tieyong and Chan, Raymond and Shan, Ying},
  booktitle=CVPR,
  pages={22139--22149},
  year={2024}
}

@article{meng2024PhyGenBench,
  title={Towards world simulator: Crafting physical commonsense-based benchmark for video generation},
  author={Meng, Fanqing and Liao, Jiaqi and Tan, Xinyu and Shao, Wenqi and Lu, Quanfeng and Zhang, Kaipeng and Cheng, Yu and Li, Dianqi and Qiao, Yu and Luo, Ping},
  journal={arXiv preprint arXiv:2410.05363},
  year={2024}
}

@misc{bansal2024videophy,
      title={{VideoPhy}-2: A Challenging Action-Centric Physical Commonsense Evaluation in Video Generation}, 
      author={Hritik Bansal and Clark Peng and Yonatan Bitton and Roman Goldenberg and Aditya Grover and Kai-Wei Chang},
      year={2025},
      eprint={2503.06800},
      archivePrefix={arXiv},
      primaryClass={cs.CV},
}

@inproceedings{xue2025phyt2v,
  title={Phyt2v: Llm-guided iterative self-refinement for physics-grounded text-to-video generation},
  author={Xue, Qiyao and Yin, Xiangyu and Yang, Boyuan and Gao, Wei},
  booktitle=CVPR,
  pages={18826--18836},
  year={2025}
}

@inproceedings{Wang2020TartanAirAD,
  title={{TartanAir}: A Dataset to Push the Limits of Visual SLAM},
  author={Wenshan Wang and Delong Zhu and Xiangwei Wang and Yaoyu Hu and Yuheng Qiu and Chen Wang and Yafei Hu and Ashish Kapoor and Sebastian A. Scherer},
  booktitle={IROS},
  year={2020},
  pages={4909-4916},
}

@inproceedings{Ling2023DL3DV10KAL,
  title={{DL3DV-10K}: A Large-Scale Scene Dataset for Deep Learning-based 3{D} Vision},
  author={Lu Ling and Yichen Sheng and Zhi Tu and Wentian Zhao and Cheng Xin and Kun Wan and Lantao Yu and Qianyu Guo and Zixun Yu and Yawen Lu and Xuanyi Li and Xingpeng Sun and Rohan Ashok and Aniruddha Mukherjee and Hao Kang and Xiangrui Kong and Gang Hua and Tianyi Zhang and Bedrich Benes and Aniket Bera},
  booktitle=CVPR,
  year={2023},
  pages={22160-22169},
}
\end{document}